\documentclass[twocolumn]{svjour3}
\smartqed  

\usepackage[utf8]{inputenc}
\usepackage{ifthen}
\usepackage{hyperref}
\usepackage{soul}
\usepackage[normalem]{ulem}
\usepackage{graphicx}
\usepackage{epsf,picinpar}
\usepackage{varioref}
\usepackage{fdsymbol}
\usepackage[dvipsnames]{xcolor}
\usepackage{enumerate}
\usepackage[framemethod=TikZ]{mdframed}
\usepackage[backend=biber,defernumbers=true,style=numeric,maxcitenames=1,maxbibnames=10]{biblatex}
\usepackage{booktabs}
\usepackage{tabularx}
\usepackage{multirow}
\usepackage{float}
\usepackage{array}
\usepackage{harveyballs}
\usepackage{makecell}
\usepackage{tablefootnote}
\usepackage[nomessages]{fp}%
\usepackage{subcaption}
\usepackage{tikz}
\usetikzlibrary{arrows}

\definecolor{barcolor}{HTML}{85d4ff}
\definecolor{subbarcolor}{HTML}{8fff85}
\definecolor{subsubbarcolor}{HTML}{d8d8d8}
\def\harveyBallsSize{0.5ex}

\newcommand{\secref}[1]{Section~\ref{#1}}

\newcommand{\figref}[1]{Figure~\ref{#1}}
\newcommand{\tabref}[1]{Table~\ref{#1}}

\newenvironment{conclusionframe}[1]
  {\mdfsetup{
    frametitle={\colorbox{white}{\space#1\space}},
    innertopmargin=-5pt,
    frametitleaboveskip=-\ht\strutbox,
    frametitlealignment=\center,
    backgroundcolor=gray!3,
    roundcorner=3pt%
    }
  \begin{mdframed}[nobreak=true]%
  }
  {\end{mdframed}}

\newcommand{\corner}{$\rotatebox[origin=c]{180}{$\Lsh$}$}
\newcommand{\included}[1]{\textbf{$\rightarrow${#1}}}

\newcolumntype{L}[1]{>{\raggedright\let\newline\\\arraybackslash\hspace{0pt}}m{#1}}
\newcolumntype{C}[1]{>{\centering\let\newline\\\arraybackslash\hspace{0pt}}m{#1}}
\newcolumntype{R}[1]{>{\raggedleft\let\newline\\\arraybackslash\hspace{0pt}}m{#1}}

\newcommand{\yes}[1][0.9ex]{%
  \harveyInternal[#1]{black}{\harveyBallFull}
}
\newcommand{\partially}[1][0.9ex]{%
  \harveyInternal[#1]{black}{\harveyBallHalf}
}

\newcommand{\harveyInternal}[3][0.9ex]{%
  \def\harveyBallsSize{#1}%
  \def\harveyBallsColor{#2}%
  \def\harveyBallsLineColor{#2}%
  #3
}
\newcommand{\inputTo}[0]{%
    \textcolor{black}{\textbf{$\xrightarrow{in}$}}
}
\newcommand{\outputFrom}[0]{%
    \textcolor{black}{\textbf{$\xleftarrow{out}$}}
}

\newcommand{\colheight}{-.25}
\newlength{\maxlen}

\newcommand{\numberofstudies}{22}

\newcommand{\maindatabar}[2][\numberofstudies]{%
    \databar[#1]{#2}{32}{barcolor}
}

\newcommand{\subdatabar}[2][\numberofstudies]{%
    \databar[#1]{#2}{36}{subbarcolor}
}

\newcommand{\subsubdatabar}[2][\numberofstudies]{%
    \databar[#1]{#2}{40}{subsubbarcolor}
}

\newcommand{\databar}[4][\numberofstudies]{%
    \settowidth{\maxlen}{#3}%
    \addtolength{\maxlen}{\tabcolsep}%
    \ifthenelse{\equal{#1}{}}{%
        \FPeval{perc}{round(100.0 / \studynumberdefault * #2, 1)}%
    }%
    {%
        \FPeval{perc}{round(100.0 / #1 * #2, 1)}%
    }%
    \FPeval\result{round(\perc / #3, 1)}%
    \rlap{\color{#4}\hspace*{-.5\tabcolsep}\rule[\colheight\ht\strutbox]{\result\maxlen}{1.2\ht\strutbox}}%
    \makebox[\dimexpr\maxlen-\tabcolsep][l]{#2 (\perc\%)}%
}

\newcommand{\subcategory}{\;\corner{}}
\newcommand{\subsubcategory}{\;\;\corner{}}

\newcommand*{\affaddr}[1]{#1} 
\newcommand*{\affmark}[1][*]{\textsuperscript{#1}}

\newcommand\xofy[3]{%
    \FPeval{perc}{round(100.0 / #2 * #1, 1)}%
    \ifthenelse{\equal{#3}{}}{%
        {#1} of {#2} (\perc\%)%
    }%
    {%
        {#1} of {#2} {#3} (\perc\%)%
    }%
}

\newcommand\xofyp[2]{%
    \FPeval{perc}{round(100.0 / #2 * #1, 1)}%
    {#1} of {#2} -- \perc\%%
}

\newcommand\perc[2]{%
    \FPeval{perc}{round(100.0 / #2 * #1, 1)}%
    \perc%
}
\journalname{Software and System Modeling}

\DeclareFieldFormat{labelnumberwidth}{#1\adddot}
\DeclareSourcemap{
  \maps[datatype=bibtex, overwrite]{
    \map{
      \perdatasource{bib/primary-studies.bib}
      \step[fieldset=KEYWORDS, fieldvalue=primary, append]
    }
  }
}

\usepackage[pscoord]{eso-pic}
\definecolor{linkblue}{RGB}{4,0,237}
\newcommand{\hreffinternal}[3]{\href{#1}{\textcolor{#3}{#2}}}
\newcommand{\hreff}[2]{\hreffinternal{#1}{#2}{linkblue}}

\newcommand{\placetextbox}[3]{
  \setbox0=\hbox{#3}
  \AddToShipoutPictureFG*{
    \put(\LenToUnit{#1\paperwidth},\LenToUnit{#2\paperheight}){\vtop{{\null}\makebox[0pt][c]{#3}}}%
  }%
}%

\begin{document}

\placetextbox{0.5}{0.99}{\large\colorbox{gray!3}{\textcolor{WildStrawberry}{\textbf{Author pre-print.}}}}%

\placetextbox{0.5}{0.97}{\large\colorbox{gray!3}{\textcolor{WildStrawberry}{Final publication: \hreff{https://doi.org/10.1007/s10270-025-01306-0}{https://doi.org/10.1007/s10270-025-01306-0}.}}}%

\placetextbox{0.5}{0.05}{\colorbox{gray!3}{\textcolor{WildStrawberry}{Author pre-print. Final publication appears in} \hreff{https://www.sosym.org/}{SoSyM}.}}%

\title{AI Simulation by Digital Twins: Systematic Survey, Reference Framework, and Mapping to a Standardized Architecture}

\titlerunning{AI Simulation by Digital Twins}

\author{
    Xiaoran Liu\protect\affmark[1] \and
    Istvan David\protect\affmark[1,2] 
}

\authorrunning{X. Liu and I. David}

\institute{
    \affaddr{\affmark[1]McMaster University, Hamilton, Canada}\\
    \affaddr{\affmark[2]McMaster Centre for Software Certification (McSCert), Hamilton, Canada}\\
    \email{liu2706@mcmaster.ca, istvan.david@mcmaster.ca}
}

\date{Received: date / Accepted: date}

\emergencystretch 3em

\maketitle

\begin{acknowledgements}
We acknowledge the support of the Natural Sciences and Engineering Research Council of Canada (NSERC), DGECR-2024-00293.
\end{acknowledgements}

\begin{abstract}

Insufficient data volume and quality are particularly pressing challenges in the adoption of modern subsymbolic AI. To alleviate these challenges, AI simulation uses virtual training environments in which AI agents can be safely and efficiently developed with simulated, synthetic data.
Digital twins open new avenues in AI simulation, as these high-fidelity virtual replicas of physical systems are equipped with state-of-the-art simulators and the ability to further interact with the physical system for additional data collection.
In this article, we report on our systematic survey of digital twin-enabled AI simulation. By analyzing 22 primary studies, we identify technological trends and derive a reference framework to situate digital twins and AI components. Based on our findings, we derive a reference framework and provide architectural guidelines by mapping it onto the ISO 23247 reference architecture for digital twins. Finally, we identify challenges and research opportunities for prospective researchers.
\end{abstract}


\section{Introduction}\label{sec:intro}

Modern artificial intelligence (AI) is enabled by massive volumes of data processed by powerful computational methods~\cite{zhou2017machine}. This is a stark contrast with traditional AI, which is supported by symbolic methods and logic~\cite{smolensky1987connectionist}. The volume and quality of available data to train AI is the cornerstone of success in modern AI. However, accessing and harvesting real-world data is a substantial barrier due to its scarcity, cost, or difficult accessibility, hindering the development of precise and resilient AI models. For example, in manufacturing, proprietary data, data silos, and sensitive operational procedures complicate the acquisition of data~\cite{farahani2023smart}. Data-related barriers, in turn, limit the applicability of otherwise powerful AI methods.

AI simulation is a prime candidate for alleviating these problems. As defined by Gartner recently, AI simulation is the technique of ``\textit{the combined application of AI and simulation technologies to jointly develop AI agents and the simulated environments in which they can be trained, tested and sometimes deployed. It includes both the use of AI to make simulations more efficient and useful, and the use of a wide range of simulation models to develop more versatile and adaptive AI systems}''~\cite{aisim-gartner}.
After modeling the phenomenon or system at hand, a simulation of the model computes the dynamic input/output behavior~\cite{vangheluwe2002introduction}, representative of the system. A simulation produces data, called the simulation trace, that represents the behavior of the simulated system over time. These traces can be used as training data for AI agents, assuming that the simulation is a faithful, valid and detailed representation of the modeled system, and that the simulation can still be executed efficiently and in a timely manner.

Digital twins (DT)~\cite{kritzinger2018digital} align well with the ambitions of AI simulation for two reasons.
First, \textbf{the emergence of DTs elevated the quality, fidelity, faithfulness, and performance of simulators}. Simulators are first-class components of DTs~\cite{david2024automated} and enablers of sophisticated services, e.g., real-time adaptation~\cite{tomin2020development}, predictive analytics~\cite{paredis2024coock}, and process control in manufacturing~\cite{bibow2020model}. These advanced services require well-performing and high-fidelity simulators---the types of simulators that align well with the goals of AI simulation.
Second, through the close coupling of the digital and physical side, \textbf{DTs allow for purposeful experimentation} with the physical twin. In practical situations, simulators might not be able to generate valid data for every request from the AI agent. This might be due to the lack of knowledge encoded in the simulator about the particular query, or outdated knowledge encoded in the simulator. In both cases, observing the real, physical setting to collect samples and update the simulation model offers an apt solution. Moreover, the closed control in DTs allows for \textit{purposeful} experimentation with the physical system rather than mere passive observation. Experimentation is the basis of simulator engineering~\cite{zeigler2018theory} and DTs allow for more targeted and automated experimentation. Thus, even in cases when a simulator might not be able to provide valid training data, the DT which the simulator is part of, might be able to compensate for missing data.

Therefore, it is plausible to anticipate that the next generation of AI simulation techniques will be heavily influenced by the further advancements of DT technology~\cite{hu2024how,qi2018digital}. The prevalence of this view has been demonstrated in a recent interview study among nineteen academic and industry participants by \textcite{muctadir2024current}, who mention that ``\textit{machine learning and reinforcement learning could possibly be combined with DTs in the future, to help to learn about complex systems (i.e., safety-critical systems) in a virtual environment, when this is difficult to do on the real-world system}.''
Similar ambitions have been identified by \textcite{mihai2022digital} as future prospects of DTs.

Anticipating such a convergence between DT, simulation, and AI technology, it is important to understand the state of affairs in digital twinning for AI simulation purposes, to identify related challenges, and to set a targeted research agenda.
This work marks a step towards converging AI simulation and DT technology. We review the state of the art on AI simulation by DTs, derive a framework, identify trends in system organization, AI flavors, and simulation, and outline future avenues of research.

\paragraph{Context and scope} In this work, we focus on \textbf{AI simulation \textit{by} digital twins}. We acknowledge the utility of the other direction, i.e., simulators of DTs being enabled by AI~\cite{legaard2023constructing}; however, we consider such works outside the scope of the current study.

\paragraph{Contributions} Our contributions are as follows.
\begin{itemize}
    \item We design, conduct, and report a \textbf{systematic survey} of the state of the art in AI simulation by digital twins.
    \item Based on the results of our survey, we derive a \textbf{conceptual reference framework} to integrate (i) digital twins and (ii) AI components for the purpose of AI simulation.
    \item For more actionable insights, we \textbf{map our reference framework onto the ISO 23247-2:2021 reference architecture} for digital twins.
    \item We identify \textbf{technological trends, key challenges, and research opportunities} in AI simulation by digital twins for prospective researchers.
\end{itemize}

\paragraph{Replicability} For independent verification, we publish a replication package containing the data and analysis scripts of our study.\footnote{The replication package of this extended version of our study is available at \url{https://doi.org/10.5281/zenodo.15602825}.
The replication package of the conference publication~\cite{liu2024ai} is available at \url{https://doi.org/10.5281/zenodo.13293238}.}

\paragraph{Novelty statement}
This paper is an extended version of our conference paper~\cite{liu2024ai} accepted for the 1st International Conference on Engineering Digital Twins (EDTconf 2024). It extends the original paper by
\begin{itemize}
    \item (i) a \textbf{new section} on mapping the DT4AI reference framework on the ISO 23247-2:2021 reference architecture for digital twins (\secref{sec:iso-mapping});
    
    \item (ii) a \textbf{new research question}, mapping, and discussion about the \textbf{technological choices} in digital twin-enabled AI simulation (\secref{sec:rq6});

    \item (iii) \textbf{more detailed analysis of the AI/ML tasks} AI simulation is used for (\secref{sec:rq3}, particularly related to \tabref{tab:activities-ai});
    
    \item (iv) \textbf{extended background section} elaborating on the key concepts of data augmentation (\secref{sec:background-dataaugmentation}) and the sim-to-real transfer (\secref{sec:background-sim2real});
    
    \item (v) \textbf{extended discussion} of the results (\secref{sec:discussion-techchoices}).
\end{itemize}

\paragraph{Structure} The rest of this article is structured as follows.
In \secref{sec:background}, we review the background topics of our work and the related work.
In \secref{sec:study-design}, we design a study to survey the state of the art in AI simulation by digital twins.
In \secref{sec:framework}, we define a conceptual reference framework for AI simulation by digital twins.
In \secref{sec:results}, we report the results of our survey.
In \secref{sec:discussion}, we discuss the results and identify open challenges and research opportunities of prospective researchers.
In \secref{sec:iso-mapping}, we map our conceptual reference framework for DT-enabled AI simulation onto the ISO 23247-2:2021 reference architecture for digital twins.
Finally, in \secref{sec:conclusion}, we draw the conclusion and identify future work.
\section{Background and Related Work}\label{sec:background}

We now review the background and the related work.

\subsection{Data challenges in AI training}

The data-related challenges of modern AI are well-documented.
In their review of fifteen key challenges in AI, \textcite[Challenge 13]{hagendorff202015} identify the problem of the acute scarcity of labels despite labeled data being a hard precondition to many AI systems.
Obtaining data of sufficient quantity and quality can be challenging. Data quality directly affects the effectiveness of model training. Common data quality issues include missing data, inconsistencies, duplications, and noise. Obtaining high-quality data typically requires data cleaning and pre-processing.
\textcite{hagendorff202015} consider these challenges ephemeral, i.e., technological advancement is expected to solve these challenges in the short run.
{One of such technological advancements to combat data scarcity is digital twin technology, which enables targeted data acquisition from physical systems and, by that, facilitates the derivation and maintenance of faithful simulators that produce realistic training data.

The improvement of data quality is mainly realized through data cleaning and preprocessing, including methods such as removing duplicates, handling missing values, and eliminating noise~\cite{garcia2016big}. In addition, automated tools and algorithms can be utilized to assess and monitor data quality, and detect and fix problems in time~\cite{ehrlinger2022survey}.
Crowdsourcing platforms (e.g., Amazon's Mechanical Turk~\cite{turk2012amazon}) can also be used for large-scale collaborative annotation to augment semi-automated tools and algorithms~\cite{neveol2011semi}.
Other alternatives, such as assisted human labeling~\cite{ashktorab2021ai,desmond2021increasing}, and labeling with ChatGPT~\cite{nguyen2024human,syriani2023assessing,syriani2024screening}, are actively researched currently.

In this work, we draw attention to the emerging topic of AI simulation as a potential solution to these problems.

\subsection{Data augmentation}\label{sec:background-dataaugmentation}

Data augmentation aims to expand the training dataset by applying transformations to the original data, including modifying already available data or generating new data points~\cite{shorten2019survey}. The primary objective of data augmentation is to enhance the quantity, quality, and variety of training data, e.g., avoid overfitting or improve the model's robustness~\cite{mumuni2022data,ko2015audio}.

Data augmentation techniques play a vital role in improving models' generalization capabilities, with applications in domains, such as audio processing~\cite{abayomi2022data}, text processing~\cite{bayer2022survey}, and image processing~\cite{shorten2019survey}. In computer vision, data augmentation methods typically include operations such as rotating, flipping, scaling, and cropping images~\cite{shorten2019survey}; in natural language processing (NLP) field, datasets can be expanded by word replacement, randomly inserting or deleting words, and reorganizing the sentence structure~\cite{bayer2022survey}. Recent advancements in deep learning techniques gave rise to more intricate data augmentation approaches, based on, e.g., Generative Adversarial Networks (GANs)~\cite{creswell2018generative}.

Although data augmentation techniques have improved significantly recently, limitations still persist. First, most methods rely on simple predefined transformations, making it challenging to simulate complex changes in real-world scenarios. Second, the quality of the augmented data is difficult to verify and may result in invalid data. Third, application methods vary across domains and most existing techniques are designed for specific tasks, limiting generalizability~\cite{abayomi2022data}~\cite{shorten2019survey}.

AI simulation is positioned as a key candidate to overcome these challenges thanks to its ability to generate highly diverse data that is more faithful to real-world settings.

\subsection{Simulation}

Simulators are programs that encode the probabilistic mechanism that represents the real phenomenon and enact this probabilistic mechanism over a sufficiently long period of time to produce simulation traces describing the real system~\cite{zeigler2018theory}.

From the '60s, computer simulation was employed in select domains by few experts until, in the '80s, it became a key enabler in solving complex engineering problems. In the past decade, advancements in digital technology shifted the typical role of simulators again, this time down to the operational phase of systems~\cite{boschert2016digital}. As a prime exemplification of this trend, simulators are first-class components of DTs~\cite{david2024automated} and enablers of the sophisticated features and services DTs provide, e.g., providing a learning environment for training purposes of human and computer agents~\cite{jaensch2018digital}.

At the core of the simulator, the physical asset is represented by a model, from which complex algorithms calculate the metrics of interest. This model captures the essential properties of the simulated asset in appropriate detail to consider the results of the simulation representative.
The execution of a simulation produces a \textit{simulation trace}, that represents the behavior of the simulated system over time~\cite{ross2022simulation}.
These simulation traces are the \textit{data} that can be used to train and tune AI agents.

\subsection{Sim-to-real}\label{sec:background-sim2real}

Sim-to-real is the procedure of transferring knowledge gained in simulated settings to real-world applications~\cite{hu2024how}.
Simulated environments provide a controlled, safe, and low-cost training and testing platform compared to real-life environments. However, simulation environments are frequently built on idealized assumptions that may ignore seemingly minor but significant real-world factors such as object friction, air resistance, light changes, noise, and sensor delays, etc.~\cite{chebotar2019closing}.
Bridging the gap between simulated and real worlds---commonly referred to as the ``reality gap''~\cite{kadian2020sim2real}---is the focus of sim-to-real transfer techniques. The lack of sim-to-real considerations makes it challenging for AI agents to match their simulation-based performance in real-world settings.

Sim-to-real plays an important role in narrowing the reality gap by ensuring that simulations produce more realistic data, a critical component for AI simulation.
While sim-to-real specializes in addressing the challenges of transferring knowledge from simulation environments to real-world settings, AI simulation covers a much broader lifecycle, including data collection, AI training, and control mechanisms.

The most common approach for sim-to-real transfer is domain randomization, with other prominent methods including knowledge distillation, domain adaption, and meta-reinforcement learning~\cite{zhao2020sim}. By enhancing the model's robustness, adaptability, and generalizability, each of these techniques helps to facilitate a better sim-to-real transfer. Utilizing these methods, sim-to-real transfer has been demonstrated in a variety of fields, including robotics~\cite{tan2018sim} and autonomous driving~\cite{hu2024how}. In order to bridge the distribution gap between the generated output and the target data, \textcite{kar2019meta} propose Meta-Sim, a sim-to-real learning model, to generate synthetic driving scenarios for the automated construction of labeled datasets relevant to downstream tasks.

\subsection{Related work}

Although our work marks the first survey on AI simulation by DTs, the benefits of combining DTs and AI have been recognized before.
In their review of applying AI in Industry 4.0, \textcite{baduge2022artificial} identify the integration potential of AI with DTs to enhance the precision of DT models and iteratively refine these models using continuously gathered data. \textcite{emmert_streib2023what} investigate techniques that combine AI and DTs, and identify ``generative modeling'', roughly analogous to AI simulation, as an opportunity with elevated potential. This underlines the importance of our work.

A related body of knowledge is the one dedicated to the opposite direction of support between AI and DTs, i.e., AI for DTs.
\textcite{yitmen2021adapted} use AI to improve the creation of DT simulation models by simplifying their structure and functionality.
\textcite{david2021inference} propose a method for inferring DT simulation models through deep reinforcement learning. Their evaluation shows that DTs augmented with reinforcement learning facilities can efficiently learn from the right signals.
\textcite{neethirajan2021is} investigates the use cases and potentials of generative adversarial networks in the livestock industry to generate simulation data for the development of DTs.

Multiple secondary studies on DT practices relate to our work.
\textcite{muctadir2024current} conduct an interview study focusing on the trends in DT development, maintenance, and operation. Their interviews with 19 experts from industry and academia reveal problematic areas, such as the lack of uniform definitions, tools, techniques, and methodologies, and call for the adoption of more rigorous software engineering practices in support of the DTs' lifecycles. Our study corroborates these findings at many points, as explained later.
\textcite{mihai2022digital} survey the enabling technologies, trends, and future prospects of DTs. A key technological prospect they identify is the strong convergence of AI and DTs. Their leads are mostly complementary to our focus as they sample techniques in which machine learning ``represents the foundation of a DT''. The broader definition of AI simulation is inclusive of this direction as well.
\section{Study design}\label{sec:study-design}

In this section, we design a protocol to systematically study the literature on digital twins for AI simulation.

\subsection{Goal and research questions}

The goal of this study is to analyze the use-cases, technical characteristics, and context of digital twins, used for AI simulation.
To this end, we formulate the following research questions.

\begin{enumerate}[\bfseries{RQ}1.]
    \item \textit{In what \textbf{domains and problems} are digital twins used to support AI simulation?}

    By addressing this research question, we aim to understand the motivation for employing digital twins for AI simulation.
    
    \item \textit{What are the \textbf{technical characteristics of digital twins} used in AI simulation?}

    We aim to understand which digital twin styles are used (e.g., twin, shadow, human-in-the-loop), how DTs are architected, and which M\&S formalisms are used for AI simulation.

    \item \textit{Which \textbf{AI/ML techniques} is digital twin-enabled AI simulation used for?}

    To answer this research question, we categorize and analyze Artificial Intelligence (AI) and Machine Learning (ML) techniques for which AI simulation is used for in some of the typical AI development activities (e.g., training, validation, etc.).

    \item \textit{What \textbf{lifecycle models} are used in support of digital twin-enabled AI simulation?}

    By addressing this research question, our goal is to understand the lifecycle of AI simulation with a particular focus on the maintenance of simulators, and whether simulated data is validated in a specific step(s) along the lifecycle.

    \item \textit{What \textbf{technologies and techniques} are used in support of AI simulation by DT?}

    We aim to identify software and hardware technologies that are typically used in DT-enabled AI simulation, and characteristic techniques to solve DT- and AI-specific challenges.
    
    \item \textit{What are the \textbf{open challenges} in DT-enabled AI simulation?}

    We aim to identify challenges to which researchers in the DT and model-driven engineering communities can contribute.

\end{enumerate}

\subsection{Search and selection}

To identify relevant studies, we employ a combination of automated search, manual search and snowballing. In the following, we elaborate on this process. \tabref{tab:numbers} reports the relevant figures.

\subsubsection{Automated search}

We construct our initial search string from the topic of interest (``AI simulation'') and its explanation (``development or training of AI or ML by digital twins''~\cite{aisim-gartner}):

\begin{small}
\begin{verbatim}
("AI simulation") OR
(("digital twin*") AND
 ("train*" OR "develop*") AND
 ("AI" OR "artificial intelligence" OR
  "ML" OR "machine learning"))
\end{verbatim}
\end{small}

Experimentation with different variations of the search string yields a negligible amount of true positives and a substantial amount of false positives. This is likely because AI simulation is a new, emergent field (explains the lack of results from the second, detailed part of the search string), and the term ``AI Simulation'' might not be widely adopted in academic works just yet (explains the lack of results from the first part of the search string).

\begin{table*}
\centering
\caption{Statistics of the search and selection rounds}
\label{tab:numbers}
\begin{tabular}{@{}lrrr@{\hspace{2pt}}lr@{}}
\toprule
\multicolumn{1}{c}{\textbf{Initial search}} & \multicolumn{1}{c}{\textbf{All}} & \multicolumn{1}{c}{\textbf{Excluded}} & \multicolumn{2}{c}{\textbf{Included}} & \multicolumn{1}{c}{\textbf{$\kappa$}} \\
Automated search & 4 & 2 & 2 & & 1.000 \\
Manual search & & & 4 & &  \\
Expert knowledge & & & 4 & & \\
\textbf{Subtotal} & & & \textbf{10} & & \\[0.5em]

\multicolumn{1}{c}{\textbf{Snowballing 1}} & \multicolumn{1}{c}{\textbf{All}} & \multicolumn{1}{c}{\textbf{Excluded}} & \multicolumn{2}{c}{\textbf{Included}} & \multicolumn{1}{c}{$\kappa$} \\

Backward & 558 & 553 & 5 & (0.90\%) &  \\
Forward & 90 & 86 & 4 & (4.44\%) &  \\
\textbf{Subtotal} & 648 & 639 & \textbf{9} & \textbf{(1.39\%)} & 0.840 \\[0.5em]

\multicolumn{1}{c}{\textbf{QA 1}} & \multicolumn{1}{c}{\textbf{All}} & \multicolumn{1}{c}{\textbf{Excluded}} & \multicolumn{2}{c}{\textbf{Retained}} & \\
\textbf{Subtotal} & 19 & 7 & \included{12} & (1.82\%) & \\[0.5em]

\midrule
\multicolumn{1}{c}{\textbf{Snowballing 2}} & \multicolumn{1}{c}{\textbf{All}} & \multicolumn{1}{c}{\textbf{Excluded}} & \multicolumn{2}{c}{\textbf{Included}} & \multicolumn{1}{c}{$\kappa$} \\

Backward\footnotemark[2] & 499 & 494 & 5 & (1.00\%)  &  \\
Forward\footnotemark[3] & 618 & 601 & 17 & (2.75\%) &  \\
\textbf{Subtotal}\footnotemark[4] & 1\,117 & 1\,095 & \textbf{22} & \textbf{(1.97\%)} & 0.662 \\[0.5em]

\multicolumn{1}{c}{\textbf{QA 2}} & \multicolumn{1}{c}{\textbf{All}} & \multicolumn{1}{c}{\textbf{Excluded}} & \multicolumn{2}{c}{\textbf{Retained}} & \\
\textbf{Subtotal}\footnotemark[5] & 19 & 9 & \included{10} & (0.89\%) & \\[0.5em]

\midrule
\textbf{Final} & 1\,775 & 1\,753 & \included{22} & (1.24\%) & \\
\bottomrule
\end{tabular}
\end{table*}

\footnotetext[2]{Of the 499 backward references, 8 were selected for inspection by interpreting their citation context in the data extraction phase.}
\footnotetext[3]{Of the 618 citations, 192 were selected via a citation-based preliminary screening.}
\footnotetext[4]{$\kappa$ calculated from the 8+192=200 studies screened by both authors.}
\footnotetext[5]{After clustering, only 19 newly included studies remain in this phase.}
\setcounter{footnote}{5}

To mitigate false positives, we use a high-level search string that finds AI simulation studies explicitly labeled as such; and augment the initial result set by manual search (\secref{sec:search-manual}) and expert knowledge (\secref{sec:search-expert-knowledge}). We use the following search string to scan Scopus, Web of Science, IEEE Xplore, and ACM Digital Library:

\begin{small}
\begin{verbatim}
("AI simulation") AND
("digital twin*" OR "digital shadow*")
\end{verbatim}
\end{small}

As reported in \tabref{tab:numbers}, the automated search finds 4 primary studies. The search strings yield 4 primary studies on Scopus, 5 on Web of Science, of which 6 remain after duplicate removal, and 4 after removing two patents.
After screening, we tentatively retain \textbf{2 primary studies}, subject to further quality assessment.

\subsubsection{Manual search}\label{sec:search-manual}

Based on our expert knowledge, we identify key venues (conferences and journals) and search for potentially relevant studies in the past five years (2019--2024). Specifically,
\begin{itemize}
    \item we scan top AI conferences for studies on DTs (IJCAI\footnote{\url{https://ijcai.org}}, ICML\footnote{\url{https://icml.cc}}, NeurIPS\footnote{\url{https://neurips.cc}}, AAAI\footnote{\url{https://aaai.org}}, ICLR\footnote{\url{https://iclr.cc}}); and 
    
    \item we scan top conferences and journals in computing, software, and systems, related to DTs for studies about AI or ML (MODELS\footnote{\url{http://modelsconference.org}}; SoSyM\footnote{\url{https://sosym.org}}, JSS\footnote{\url{https://sciencedirect.com/journal/journal-of-systems-and-software}}, IEEE Software\footnote{\url{https://computer.org/csdl/magazine/so}}).
\end{itemize}

When choosing AI venues, we consider the currently top (CORE-A*) conferences in AI. Considering the conference-focused publication trends in AI, we deem this sample sufficient for our purposes.
When choosing DT venues, we rely on our expert knowledge and the publication venues of the community-curated list of key publications by the Engineering Digital Twins (EDT) Community~\cite{edt-community-papers}. The selected ones are flagship publication outlets for the DT community (including a CORE-A conference and multiple journals).

When scanning conferences, we also consider their satellite events, such as workshops. We scan the past five editions of each conference, given that AI simulation is a relatively new concept that appeared in Gartner's glossary in 2023 for the first time.

We select potentially relevant studies by checking them against the exclusion criteria (\secref{sec:excriteria}) using adaptive reading depth~\cite{petersen2008systematic}. That is, we first check the title and abstract of the study, and if deemed relevant, we assess whether the study merits consideration to be included by processing the full text.
We tentatively include \textbf{4 primary studies}, subject to further quality assessment.

\subsubsection{Expert knowledge}\label{sec:search-expert-knowledge}

To round out the initial phase of the search, we add studies that we are familiar with and have not been found by the search string or manual search. 
Similar to the manual search phase, we again select relevant studies by checking them against the exclusion criteria (\secref{sec:excriteria}) using adaptive reading depth~\cite{petersen2008systematic} (first checking the title and abstract of the study, and if relevant, scanning the full-text for details).
We tentatively include \textbf{4 primary studies}, subject to further quality assessment.

\phantom{}

After this phase, the initial set consists of \textbf{10 primary studies}, subject to further quality assessment.

\subsubsection{Snowballing}
We apply two rounds of backward and forward snowballing to enrich the corpus. The studies we include in the second round of snowballing align well with the information from already included studies with minimal new or unexpected findings. Thus, we decide to conclude snowballing after two rounds.

For backward snowballing, we extract references from primary studies manually. For forward snowballing, we follow the recommendations of \textcite{wohlin2020guidelines} and extract references from Google Scholar. We automate this step through Publish or Perish~\cite{pop-website}.

In the first snowballing round, we apply an exhaustive snowballing strategy in which both researchers screen every study.
We observe a high kappa of 0.84 (``almost perfect agreement''). We assert that the level of agreement allows for a more rapid snowballing style in subsequent snowballing rounds.
We tentatively include \textbf{9 primary studies}, subject to further quality assessment.

In the second snowballing round, we apply a more agile snowballing strategy. In backward snowballing, we follow \textcite{wohlin2014guidelines} and mark potentially relevant references as we examine studies in the data extraction phase. Excluding duplicates, we eventually mark 8 references of the total 499 as relevant. These 8 references are screened by both researchers and 5 are included. 
In forward snowballing, one researcher conducts a preliminary screening in which clearly irrelevant studies are excluded. Of the total 618 studies, 192 are retained for screening by both reviewers.
We observe a kappa of 0.662 (``substantial agreement''), which we find satisfactory considering that we mitigated the threat of kappa inflation by excluding a significant number of irrelevant studies.
We tentatively include \textbf{22 primary studies}, of which 3 are from the same group of authors we already have in our corpus, and on the same topic. Thus, we apply clustering and nominate one study from each cluster as the representative primary study. Eventually, we consider \textbf{19 primary studies} after this round, subject to further quality assessment.

After each snowballing phase, newly considered publications go through the same evaluation process as prior studies.

\subsubsection{Exclusion criteria}\label{sec:excriteria}

We use the following exclusion criteria to filter works that are not relevant to our study. We use these criteria in the manual search and the snowballing rounds.
A study is excluded if it meets at least one exclusion criterion. Exclusion criteria are evaluated based on the \textit{full reference} (title, authors, venue) and the \textit{abstract}, by both authors.

\begin{enumerate}[\bfseries{E}1.]
    \item No or unclear DT; or the DT is not used for AI simulation.
    \item No or unclear AI/ML technique.
    \item Not DT for AI -- either no link between DT and AI, or the opposite direction (AI for DT).
    \item Other: off-topic; not English; not publicly available; secondary or tertiary studies; full proceedings; short papers ($<5$ pages).
\end{enumerate}

\subsubsection{Quality assessment}\label{sec:qa}

In accordance with the guidelines of \textcite{kitchenham2007guidelines}, we define a checklist to assess the quality of the corpus. Quality criteria are derived from the research questions.
Each question is answered by ``yes'' (1 point), ``partially'' (0.5 points), or ``no'' (0 points), based on the full text. To retain a study, we require a score of at least 2/4 points (50\%).

\begin{enumerate}[\bfseries{Q}1.]
    \item Digital twinning scenario clearly described.
    \item Simulation method clearly described.
    \item AI/ML method clearly identified.
    \item Acknowledges limitations and challenges.
\end{enumerate}

After the first round of snowballing, we assess studies included in the initial round and first snowballing round. Of the total 19 candidate studies, we exclude 7 and retain \textbf{12 primary studies}. 
After the second round of snowballing, we assess studies included in the second snowballing round. Of the total 19 candidate studies, we exclude 9 and retain \textbf{10 primary studies}.

Eventually, our corpus consists of \textbf{22 primary studies}.

\subsubsection{Threats to validity and quality assessment} Here, we identify the key threats to the validity, elaborate on the mitigation strategies we applied, and assess the quality of the study.

\textit{External validity.}
External validity is concerned with the generalizability of results. Our work is focusing on AI simulation through digital twins and therefore, some takeaways cannot be safely extrapolated to AI simulation in general. We mitigated such threats by being explicit about digital twins and digital shadows in our search strings and the manual search.

\textit{Construct validity.}
Our observations are artifacts of the sampled studies. Potential selection bias and missed publications may impact our observations and threaten the construct validity of this study. To mitigate this threat, we employed a diverse selection process (automated search, manual search, and input from expert knowledge), as well as snowballing until saturation~\cite{greenhalgh2005effectiveness}.

\textit{Internal validity.}
We may have missed some works due to terminology. ``AI Simulation'' is an emerging concept. Nonetheless, our scope, which is specific to digital twins, narrows our search and provides us with a sufficiently descriptive search term that finds relevant studies.
Selection bias may be present in our work due to applying only two rounds of snowballing. However, the low inclusion rate of 0.89\% at the end of the snowballing phase suggests that additional snowballing would yield minimal value.

\paragraph{Study quality} Our work scores 72.7\% in the particularly rigorous quality checklist of \textcite{petersen2015guidelines}. (Need for review: 1 point; search strategy: 2 points; evaluation of the search: 2 points; extraction and classification: 2 points; study validity: 1 point.) This quality score is \textit{significantly} higher than the typical values in software engineering. (\textcite{petersen2015guidelines} reports a median of 33\%, with only 25\% of their sampled studies having a quality score of above 40\%.)
We consider our study of \textbf{high quality}, meeting Petersen's criteria well.

\subsection{Publication trends and quality}\label{sec:pub-trends}

\figref{fig:pubtrends} reports the basic mappings of publication trends in our corpus.

The number of publications shows an increasing trend, with a clear increase in publication output in 2023, constituting half of the corpus. The relatively low number of studies in 2024 is partly due to our study being conducted in Q2/2024 and possibly due to seasonal variations in area-specific publication trends (e.g., timing of conferences). We observe an increasing interest in AI simulation.
\xofy{15}{22}{studies} are journal articles, suggesting mature research our analysis draws on.
The high quality of the corpus is further demonstrated by the high number of top publishers.

With that said, the \textit{reporting} quality of publications (\figref{fig:qscores}) is moderate, scoring around 63.6\% overall. This is score is due to the largely ignored details about simulation formalisms and methods (Q2, 45.5\%) and the lack of broad vision about challenges and research recommendations (Q4, 43.2\%). However, AI aspects are particularly well-documented (Q3, 90.9\%), and the technical details of digital twinning are sufficiently presented (Q1, 75\%).

Overall, we judge the corpus to be in a good shape to allow for sound conclusions about digital twinning and AI within reasonable boundaries of validity, but we anticipate limited leads about simulation formalisms and methods.

\begin{figure}[t]
    \centering
    \begin{subfigure}{\linewidth}
        \includegraphics[trim = 0 0.4cm 1.4cm 0.4cm,clip,width=\linewidth]{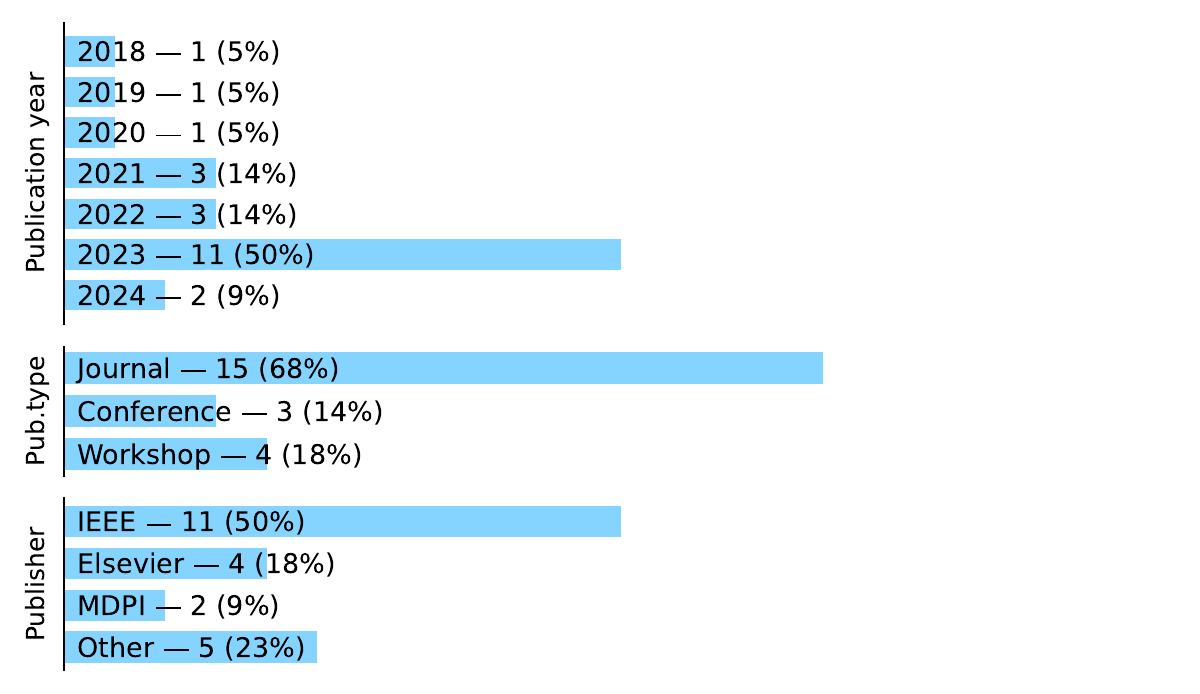}
        \label{fig:pubstats}
        \caption{Studies (as of June 2024)}
    \end{subfigure}\\[1em]
    \begin{subfigure}{\linewidth}
        \includegraphics[trim = 0 0.4cm 1.4cm 0.4cm,clip,width=\linewidth]{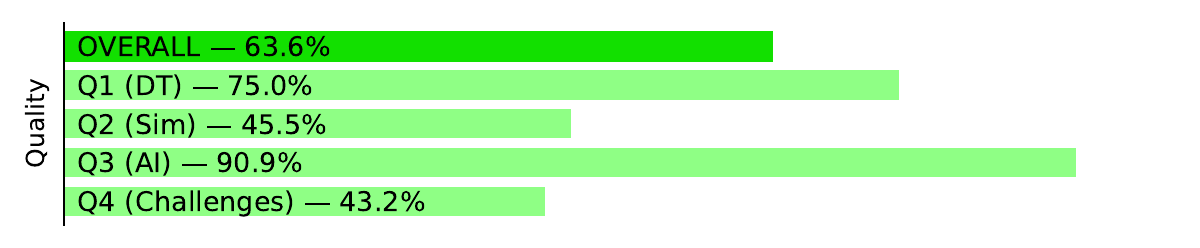}
        \caption{Quality scores}
        \label{fig:qscores}
    \end{subfigure}
    \caption{Publication trends}
    \label{fig:pubtrends}
\end{figure}

\section{The DT4AI Framework}\label{sec:framework}

To integrate DTs, AI, and simulation, we construct a conceptual reference framework from the sampled primary studies.
We rely on a mixed sample- and case-based generalization~\cite{wieringa2015six}. This approach is particularly useful when constructing middle-range theories that balance generality with practicality, such as engineering sciences.
In \secref{sec:study-design}, we sampled a statistically adequate corpus. Subsequently, we decompose each study individually into architectural units as architectural abstractions allow for better judging of similarity between cases~\cite{wieringa2015six}. Finally, we identify recurring patterns.

The resulting \textbf{DT4AI framework} is shown in \figref{fig:framework} and defines the following concepts. (The concepts are labeled in the natural order of their occurrence.)

\begin{figure}[t]
    \centering
    \includegraphics[width=0.8\linewidth]{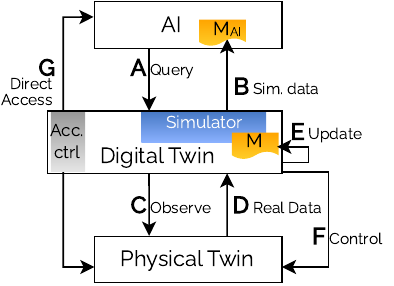}
    \caption{The DT4AI framework}
    \label{fig:framework}
\end{figure}

\begin{table*}
\centering
\caption{Variation points in the DT4AI framework}
\label{tab:framework}
\begin{tabular}{@{}ll@{}}
\toprule

\multicolumn{2}{c}{\textbf{AI training}} \\ 
\textbf{A} Query & \{Implicit, Explicit\} \\
\textbf{B} Sim. data volume & \{Big data , Small data\} \\
\textbf{A-B} Training fashion & \{Batch, Live\}\\

\midrule \multicolumn{2}{c}{\textbf{Data collection}} \\ 
\textbf{C} Observe & \{Passive observation, Active experimentation\}\\
\textbf{D} Data & \{Stationary historical data, Low-context data update, High-context data update\}\\
\textbf{C-D} Observe/Data trigger & \{Automated, On-demand\}\\
\textbf{E} Update synchronicity & \{Synchronous, Asynchronous\}\\

\midrule \multicolumn{2}{c}{\textbf{Control}} \\ 
\textbf{F} Control & \{In-place control, Deploy-and-Control\} \\

\bottomrule
\end{tabular}
\end{table*}

\subsection{Components}
The key components of the framework are the \textit{AI} agent under training, the \textit{Digital Twin} and its physical counterpart, the \textit{Physical Twin}. We follow the standard notions of digital and physical twin as defined by \textcite{kritzinger2018digital}.

For the sake of generality, we consider the \textit{AI} and \textit{Digital Twin} being separate components. This allows for offering training as a service for external AI agents, as well as training an AI agent that is part of the digital twin.

\subsection{Interactions}

\subsubsection{AI training}

AI training is an interplay between the \textit{AI} and the \textit{Digital Twin}.

\begin{description}

    \item[A: \textit{Query}.] Represents the request for data issued by the \textit{AI} component to the \textit{Digital Twin}. As shown in \tabref{tab:framework}, the \textit{Query} can be either explicit (the \textit{AI} agent actively pulling data) or implicit (the \textit{Digital Twin} pushing data).
    
    \item[B: \textit{Simulated data}.] The result of a simulation is a simulation trace, i.e., the data the \textit{AI} component receives in response to the \textit{Query}. The \textit{Digital Twin} is equipped with a (set of) model(s) \textit{M}, which serves the input to the \textit{Simulator}.
    The \textbf{A-B} training cycle can take either a batch or live format. In the former, the \textit{Trace volume} is big data; in the latter, the trace consists of small pieces of data.
\end{description}

\subsubsection{Data collection}

Data collection is an interplay between the \textit{Digital} and the \textit{Physical Twin}.

\begin{description}
    \item[C: \textit{Observe}.] The \textit{Digital Twin} is connected to the \textit{Physical Twin} through the usual data link and is able to passively observe or actively interrogate the \textit{Physical Twin} (\tabref{tab:framework}).

    \item[D: \textit{Real data}.] Represents the data collected from the \textit{Physical Twin}. Depending on the type of the \textit{Observation}, \textit{Data} might be of low context, i.e., large volumes with low information value~\cite{david2024infonomics} (in case of passive observation); or of high context, i.e., smaller volumes of data in response to active experimentation. In situations when the \textit{Digital Twin} gets detached from the \textit{Physical Twin}, e.g., due to the retirement of the latter, data can be historical as well.
    As shown in \tabref{tab:framework}, the \textbf{C-D} Observe/Data cycle can be automated (scheduled by the \textit{Digital Twin}) or on-demand (based on the requests of the \textit{AI} or human operators).
    
    \item[E: \textit{Update}.] After collecting data from the \textit{Physical Twin}, the model (\textit{M}) of the \textit{Digital Twin} needs to be updated in order to reflect the new data in simulations and transitively. This \textit{Update} can be achieved in a synchronous (blocking behavior but easier implementation) or asynchronous fashion (non-blocking behavior but more complex implementation, e.g., timeout and request obsolescence management).
\end{description}

\subsubsection{Control and access control}

Control and access control are interplays between the \textit{Digital Twin} and the \textit{Physical Twin}.

\begin{description}
    \item[F: \textit{Control}.] As customary, the \textit{Digital Twin} can control the \textit{Physical Twin} through the usual control links. As listed in \tabref{tab:framework}, control can be achieved \textit{in-place}, e.g., a learned policy on the digital side can govern the behavior of the physical system; or (parts of) the control logic can be \textit{deployed} onto the \textit{Physical Twin} for local control.

    \item[G: \textit{Access control}.] The \textit{AI} component might interact with the \textit{Physical Twin} without the participation of the simulation facilities of the \textit{Digital Twin}. In these situations, the \textit{Digital Twin} provides \textit{Access control} to the \textit{Physical Twin}. We do not consider this case alone a DT-enabled AI simulation; however, as discussed later, direct control with the \textit{Physical Twin} can be used in combination with \textit{AI} simulation, e.g., to adapt the trained agent to a physical setting.
\end{description}

The framework enables the systematic comparison and discussion of various DT-enabled AI simulation approaches.
In \secref{sec:results}, we organize evidence along the framework by instantiating it for the different flavors of architectures, AI methods, and simulation lifecycles we found in the state of the art.
\section{State of the art}\label{sec:results}

In this section, we report the key results of our empirical inquiry into the state of the art of AI simulation by DTs.
Readers are referred to the replication package for the complete data extraction sheet.

\subsection{Domains and problems (RQ1)}\label{sec:rq1}

\begin{table*}[t]
\centering
\caption{Application domains}
\label{tab:results-domains}
\begin{tabular}{@{}p{3cm}lp{12cm}@{}}
\toprule
\multicolumn{1}{c}{\textbf{Domain}} & \multicolumn{1}{c}{\textbf{\#Studies}} & \multicolumn{1}{c}{\textbf{Studies}} \\ \midrule
Networks & \maindatabar{11} & \\
\subcategory{} Wireless & \subdatabar{8} & \cite{cui2023digital,deng2021digital,dong2019deep,li2022when,li2023optimization,shui2023cell,vila2023design,zhang2023digital} \\
\subcategory{} Edge & \subdatabar{2} & \cite{guo2023intelligent,liu2022digital} \\
\subcategory{} General & \subdatabar{1} & \cite{hammar2023digital} \\

Robotics and AVs & \maindatabar{6} & \cite{matulis2021robot,shen2022deep,tang2023digital,verner2018robot,yang2024social,zhang2024automated} \\

Manufacturing & \maindatabar{2} & \cite{alexopoulos2020digital,xia2021digital} \\

Energy & \maindatabar{1} & \cite{tubeuf2023increasing} \\

Urban & \maindatabar{1} & \cite{pun2023neural} \\

Agriculture & \maindatabar{1} & \cite{david2023digital} \\

\bottomrule
\end{tabular}
\end{table*}

\begin{table*}[t]
\centering
\caption{Architectural choices}
\label{tab:patterns-arch}
\begin{tabular}{@{}p{3cm}lp{12cm}@{}}
\toprule
\multicolumn{1}{c}{\textbf{Architecture}} & \multicolumn{1}{c}{\textbf{\#Studies}} & \multicolumn{1}{c}{\textbf{Studies}} \\ \midrule

Digital twin & \maindatabar{19} & \\

\subcategory{} Autonomous & \subdatabar{16} & \cite{alexopoulos2020digital,dong2019deep,cui2023digital,guo2023intelligent,hammar2023digital,li2022when,liu2022digital,matulis2021robot,deng2021digital,shui2023cell,tang2023digital,tubeuf2023increasing,vila2023design,xia2021digital,zhang2024automated,zhang2023digital} \\

\subcategory{} Human-supervised & \subdatabar{2} & \cite{verner2018robot,yang2024social} \\

\subcategory{} Human-actuated & \subdatabar{1} & \cite{david2023digital} \\

Digital shadow & \maindatabar{2} & \cite{shen2022deep,li2023optimization} \\

Digital model & \maindatabar{1} & \cite{pun2023neural} \\

Policy DT & \maindatabar{1} & \cite{tubeuf2023increasing} \\


\bottomrule
\end{tabular}
\end{table*}

As shown in \tabref{tab:results-domains}, half of the primary studies we sampled focus on a network problem.
Wireless networks (\xofyp{8}{22}) are the most represented, typically focusing on various optimization tasks by machine learning, such as optimization of resource allocation in 5G+ networks~\cite{zhang2023digital} and edge computing~\cite{dong2019deep}.
Robotics, including the management of automated vehicles (AVs) accounts for \xofy{6}{22}{} cases, with typical examples of training AI models for the control of ordinary~\cite{matulis2021robot} and underwater~\cite{yang2024social} robot arms, and controlling the flocking motion of unmanned aerial vehicles (UAVs)~\cite{shen2022deep}.

The common trait of addressed problems is their high complexity (e.g., control in dense fluid dynamics~\cite{yang2024social}) and sparse data from real observations (e.g., in slowly changing settings of agriculture~\cite{david2023digital}).

\phantom{}

\begin{conclusionframe}{RQ1: Domains and problems}
AI simulation is primarily used in problems with \textit{high complexity} and \textit{sparse or inaccessible data} from real observations. Networks and robotics are the most prominent adoption domains, accounting for over three-quarters of sampled studies.
\end{conclusionframe}

\subsection{Technical characteristics of DTs (RQ2)}\label{sec:rq2}

To analyze the technical characteristics of digital twins used in AI simulation, we rely on the superset of taxonomies by \textcite{kritzinger2018digital} and \textcite{david2024infonomics} as our initial values. The former defines the foundational classes of digital model, digital shadow, and digital twin; the latter extends this classification by defining human-supervised and human-actuated digital twins situated between fully autonomous digital twins and non-autonomous digital shadows.

As shown in \tabref{tab:patterns-arch}, the majority of the sampled AI simulation techniques (\xofyp{19}{22}) implement a digital twin. The corresponding architecture is shown in \figref{fig:patterns-arch-dt} as an instantiation of the DT4AI framework. Most of these techniques (\xofyp{16}{22} overall) implement fully autonomous digital twins, and only a fraction relies on human supervision~\cite{verner2018robot} or human actuation~\cite{david2023digital}.
The rest of the architectural patterns in \figref{fig:patterns-arch} are seldom encountered. Digital shadows and models account for only \xofy{3}{22}{} studies.

\begin{figure*}[t]
    \centering    
    \begin{subfigure}{0.24\textwidth}
        \includegraphics[height=4.25cm]{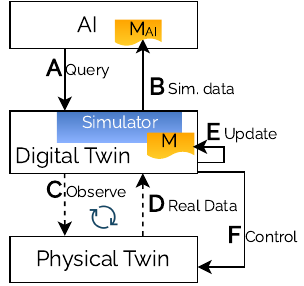}
        \caption{Experimentable DT}
        \label{fig:patterns-arch-dt}
    \end{subfigure}
    \hfill
    \begin{subfigure}{0.24\textwidth}
        \includegraphics[height=4.25cm]{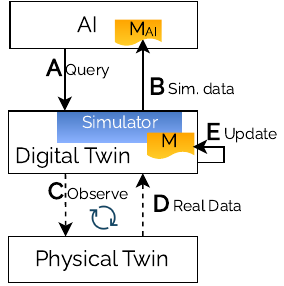}
        \caption{Experimentable DS}
        \label{fig:patterns-arch-ds}
    \end{subfigure}
    \hfill
    \begin{subfigure}{0.24\textwidth}
        \includegraphics[height=4.25cm]{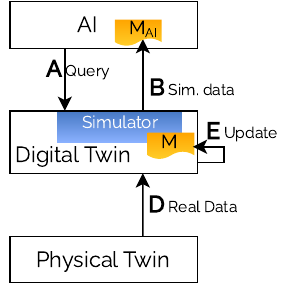}
        \caption{Experimentable Model}
        \label{fig:patterns-arch-dm}
    \end{subfigure}
    \hfill
    \begin{subfigure}{0.24\textwidth}
        \hspace{-8mm}
        \includegraphics[height=4.25cm]{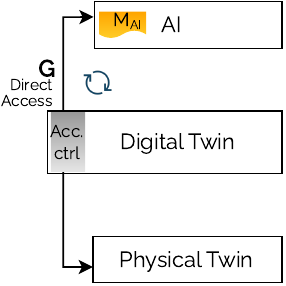}
        \caption{Policy DT}
        \label{fig:patterns-arch-policy}
    \end{subfigure}
    \caption{RQ2: Architectural patterns}
    \label{fig:patterns-arch}
\end{figure*}

\begin{table*}[t]
\centering
\caption{Modeling and simulation formalisms}
\label{tab:patterns-formalisms}
\begin{tabular}{@{}p{3cm}lp{12cm}@{}}
\toprule
\multicolumn{1}{c}{\textbf{Formalism}} & \multicolumn{1}{c}{\textbf{\#Studies}} & \multicolumn{1}{c}{\textbf{Studies}} \\ \midrule

Network models & \maindatabar{7} & \cite{cui2023digital,deng2021digital,dong2019deep,guo2023intelligent,liu2022digital,shui2023cell,vila2023design} \\

Physics & \maindatabar{5} & \cite{david2023digital,li2022when,xia2021digital,yang2024social,zhang2024automated} \\

CAD, Geometry & \maindatabar{5} & \cite{alexopoulos2020digital,shen2022deep,verner2018robot,xia2021digital,zhang2024automated} \\

Process models & \maindatabar{3} & \cite{dong2019deep,hammar2023digital,tubeuf2023increasing} \\

DEVS & \maindatabar{1} & \cite{david2023digital} \\

Unclear (DNNs, etc) & \maindatabar{5} & \cite{li2023optimization,matulis2021robot,pun2023neural,tang2023digital,zhang2023digital} \\


\bottomrule
\end{tabular}
\end{table*}

\begin{table*}[t]
\centering
\caption{DT architecture standards or reference frameworks}
\label{tab:patterns-standards}
\begin{tabular}{@{}p{3cm}lp{12cm}@{}}
\toprule
\multicolumn{1}{c}{\textbf{Standard}} & \multicolumn{1}{c}{\textbf{\#Studies}} & \multicolumn{1}{c}{\textbf{Studies}} \\ \midrule

No standard & \maindatabar{21} & \cite{cui2023digital,david2023digital,deng2021digital,dong2019deep,guo2023intelligent,hammar2023digital,li2022when,liu2022digital,li2023optimization,matulis2021robot,pun2023neural,shen2022deep,shui2023cell,tang2023digital,tubeuf2023increasing,verner2018robot,vila2023design,xia2021digital,yang2024social,zhang2024automated,zhang2023digital} \\

RAMI4.0 & \maindatabar{1} & \cite{alexopoulos2020digital} \\
\bottomrule
\end{tabular}
\end{table*}

The instantiation of the DT classes of \textcite{kritzinger2018digital} is shown in \figref{fig:patterns-arch}. \textit{Experimentable Digital Twin}s (\figref{fig:patterns-arch-dt}) and \textit{Experimentable Digital Shadow}s (\figref{fig:patterns-arch-ds}) implement the \textbf{C-D} observation loop in an asynchronous way (dashed arrows). This is contrasted with the synchronous input in \textit{Experimentable Model}s (\figref{fig:patterns-arch-ds}). We observed one case in which a DT is used as a proxy to the physical system for the AI agent to interact with~\cite{tubeuf2023increasing}. Here, the DT acts as a policy enforcer, hence the name \textit{Policy Digital Twin}. However, this pattern only appears in combination with a full DT.

\tabref{tab:patterns-formalisms} summarizes the simulation formalisms in the sampled studies. We mostly observe network models, e.g., channel state information~\cite{deng2021digital} and topology models~\cite{liu2022digital} (\xofyp{7}{22}); models of physics~\cite{yang2024social,li2022when} (\xofyp{5}{22}); and models of geometry and CAx models, e.g., CAD~\cite{zhang2024automated} and CAM/CAE~\cite{alexopoulos2020digital} (\xofyp{5}{22}). This aligns with the high number of network problems (\secref{sec:rq1}). Works that use AI models to encode the simulation model generally do not report the modeling formalism the AI model encodes.

As shown in \tabref{tab:patterns-standards}, DT architectural standards or reference frameworks are seldom used. We found one study with a standardized architecture (RAMI4.0 by \textcite{alexopoulos2020digital}).

\phantom{}

\begin{conclusionframe}{RQ2: Digital Twins}
AI simulation chiefly runs through \textit{genuine digital twins} of the \textit{autonomous} kind, but standardization is lagging behind.
\end{conclusionframe}

\subsection{AI and ML techniques (RQ3)}\label{sec:rq3}

As shown in \tabref{tab:patterns-ai}, the majority of the sampled AI simulation techniques (\xofyp{18}{22}) rely on some form of reinforcement learning. Deep Reinforcement Learning (DRL, \xofyp{13}{22}) is a heavily favored choice, with more value-based methods (\xofyp{8}{22}) than policy-based (\xofyp{5}{22}) ones. A deeper look into the details reveals state-of-the-art AI algorithms. Among value-based deep reinforcement learning, we typically find variants of Deep Q Networks~\cite{tang2023digital}; in policy-based methods, we find algorithms such as proximal policy optimization~\cite {matulis2021robot} and deep deterministic policy gradient~\cite{li2022when}. The choice of AI methods is rounded out by some approaches adopting deep learning (DL, \xofyp{4}{22}; e.g., \cite{pun2023neural}) and one case of transfer learning (TL, \xofyp{1}{22}; e.g., \cite{tubeuf2023increasing}).

\begin{figure*}
    \centering    
    \begin{subfigure}{0.32\textwidth}
        \includegraphics[height=4.25cm]{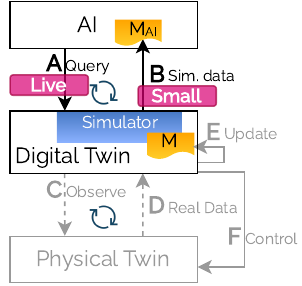}
        \caption{Reinforcement Learning}
        \label{fig:patterns-ai-rl}
    \end{subfigure}
    \hfill
    \begin{subfigure}{0.32\textwidth}
        \includegraphics[height=4.25cm]{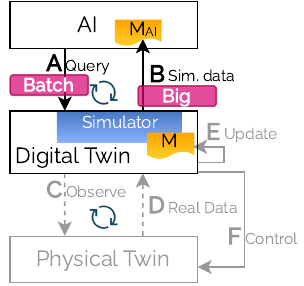}
        \caption{Deep Learning}
        \label{fig:patterns-ai-dl}
    \end{subfigure}
    \hfill
    \begin{subfigure}{0.34\textwidth}
        \includegraphics[height=4.25cm]{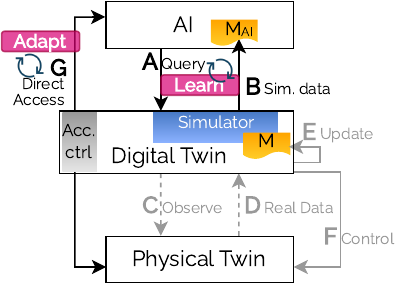}
        \caption{Transfer Learning}
        \label{fig:patterns-ai-tl}
    \end{subfigure}
    \caption{RQ3: AI patterns (relevant components highlighted)}
    \label{fig:patterns-ai}
\end{figure*}

\begin{table*}[t]
\centering
\caption{AI methods}
\label{tab:patterns-ai}
\begin{tabular}{@{}p{3cm}lp{12cm}@{}}
\toprule
\multicolumn{1}{c}{\textbf{AI method}} & \multicolumn{1}{c}{\textbf{\#Studies}} & \multicolumn{1}{c}{\textbf{Studies}} \\ \midrule

RL & \maindatabar{18} &  \\
\subcategory{} DRL & \subdatabar{13} & \\
\subsubcategory{} Value & \subsubdatabar{8} & \cite{cui2023digital,liu2022digital,shui2023cell,tang2023digital,vila2023design,xia2021digital,zhang2024automated,zhang2023digital} \\
\subsubcategory{} Policy & \subsubdatabar{5} & \cite{guo2023intelligent,li2022when,li2023optimization,matulis2021robot,shen2022deep} \\

\subcategory{} Vanilla & \subdatabar{5} & \cite{deng2021digital,hammar2023digital,tubeuf2023increasing,verner2018robot,yang2024social} \\

DL & \maindatabar{4} & \cite{alexopoulos2020digital,david2023digital,dong2019deep,pun2023neural} \\

TL & \maindatabar{1} & \cite{tubeuf2023increasing} \\
\bottomrule
\end{tabular}
\end{table*}

\begin{table*}[t]

\centering
\caption{AI/ML activities}
\label{tab:activities-ai}
\begin{tabular}{@{}p{5cm}lp{10cm}@{}}
\toprule
\multicolumn{1}{c}{\textbf{AI}} & \multicolumn{1}{c}{\textbf{\#Studies}} & \multicolumn{1}{c}{\textbf{Studies}} \\ \midrule

Virtual training environment & \maindatabar{14} & \cite{cui2023digital,deng2021digital,guo2023intelligent,hammar2023digital,li2023optimization,liu2022digital,shen2022deep,tang2023digital,verner2018robot,vila2023design,xia2021digital,yang2024social,zhang2024automated,zhang2023digital} \\

Dataset generation/labeling & \maindatabar{8} & \cite{alexopoulos2020digital,david2023digital,dong2019deep,li2022when,matulis2021robot,pun2023neural,shui2023cell,tubeuf2023increasing}  \\

\bottomrule
\end{tabular}


\end{table*}

The corresponding instantiations of the DT4AI framework are shown in \figref{fig:patterns-ai}. Structurally, \textit{Reinforcement learning} (\figref{fig:patterns-ai-rl}) and \textit{Deep learning} (\figref{fig:patterns-ai-rl}) are identical. However, there are important differences in the interactions within the \textbf{A-B} learning cycle. \textit{Reinforcement learning} establishes a \textit{live} interaction, where the \textit{AI} issues frequent, short queries for \textit{small} amounts of simulated data. In contrast, in \textit{Deep learning} uses infrequent queries (often a singular one) to which the \textit{Digital twin} responds with \textit{big} data. \textit{Transfer learning} makes use of the \textit{Physical twin}, for which the AI agent uses the \textit{Policy DT} pattern discussed in \secref{sec:rq2}. After the \textit{learning} phase, the \textit{AI} interacts with the \textit{Physical twin} to \textit{adapt} the previously learned knowledge---either to adopt the knowledge to a changing environment or to mitigate sim-to-real threats~\cite{zhao2020sim}. In support of this process, the \textit{Digital Twin} ensures the necessary reliability, safety, and security measures~\cite{tubeuf2023increasing}.

\tabref{tab:activities-ai} summarizes the specific AI simulation activities we identified in the sampled primary studies. \xofyp{14}{22} studies consider AI simulation as a \textit{virtual training environment} for AI agents, and develop training facilities accordingly. For example, \textcite{shen2022deep} use a digital twin as a safe and cost-effective training environment for unmanned aerial vehicles (UAVs), with a deep reinforcement learning (DRL) module keeping the simulation model within the digital twin up-to-date.
\textcite{verner2018robot} use a digital twin to provide robots with a safe experimentation environment to rapidly learn motion responses to previously unseen situations.
\xofyp{8}{22} studies specifically focus on the task of \textit{data generation or labeling}. For example, \textcite{alexopoulos2020digital} support the generation and labeling of virtually created datasets for manufacturing AI agents. \textcite{pun2023neural} augment images captured by cars to generate different lighting scenarios for training computer vision agents.

While AI training in virtual training environments is seen as an online training method, dataset generation is more on the offline end. That is, studies that consider AI simulation in a virtual training environment, often rely on frequent interactions between the AI agent and the training environment (i.e., the digital twin); while those relying on data generation and labeling, consider a more simplified way of communication between the AI agent and the digital twin.

\phantom{}

\begin{conclusionframe}{RQ3: AI/ML techniques}
AI simulation is predominantly used for \textit{training} purposes of \textit{reinforcement learning} agents, especially in combination with \textit{deep learning} (deep reinforcement learning).
\end{conclusionframe}

\subsection{Simulator lifecycle models (RQ4)}\label{sec:rq4}

\begin{figure*}[t]
    \centering    
    \begin{subfigure}{0.32\textwidth}
        \includegraphics[height=4.25cm]{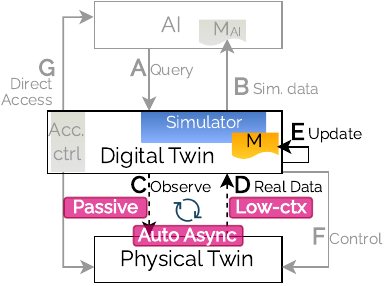}
        \caption{Automated async. update}
        \label{fig:patterns-sim-auto}
    \end{subfigure}
    \hfill
    \begin{subfigure}{0.32\textwidth}
        \includegraphics[height=4.25cm]{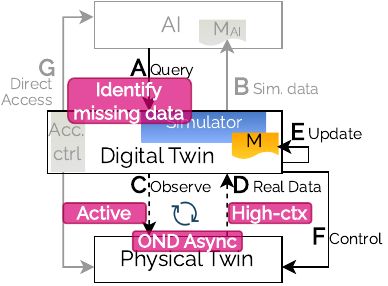}
        \caption{On-demand async. update}
        \label{fig:patterns-sim-ond-async}
    \end{subfigure}
    \hfill
    \begin{subfigure}{0.32\textwidth}
        \includegraphics[height=4.25cm]{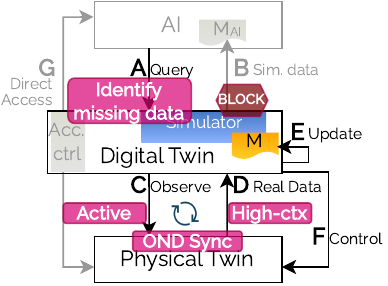}
        \caption{On-demand sync. update}
        \label{fig:patterns-sim-ond-sync}
    \end{subfigure}
    \caption{RQ4: Simulator lifecycle patterns (relevant components highlighted)}
    \label{fig:patterns-sim}
\end{figure*}

\begin{table*}[t]
\centering
\caption{Simulator maintenance patterns}
\label{tab:patterns-sim}
\begin{tabular}{@{}p{3cm}lp{12cm}@{}}
\toprule
\multicolumn{1}{c}{\textbf{Update}} & \multicolumn{1}{c}{\textbf{\#Studies}} & \multicolumn{1}{c}{\textbf{Studies}} \\ \midrule
Automated & \maindatabar{11} & \cite{cui2023digital,deng2021digital,dong2019deep,guo2023intelligent,liu2022digital,matulis2021robot,shen2022deep,shui2023cell,tang2023digital,zhang2024automated,zhang2023digital}  \\
On-demand & \maindatabar{2} & \cite{alexopoulos2020digital,vila2023design} \\
No update & \maindatabar{9} & \cite{david2023digital,hammar2023digital,li2022when,li2023optimization,pun2023neural,tubeuf2023increasing,verner2018robot,xia2021digital,yang2024social} \\
\bottomrule
\end{tabular}
\end{table*}

We aim to understand the lifecycle models along which digital twins and, in particular, simulator components are being used and maintained. Unfortunately, the low attention to detail in the simulations aspect (see \figref{fig:pubtrends}) makes it challenging to derive in-depth insights.

In general, we observe that AI simulation is provided as a service by the digital twin, and there is \textit{no need to detach} the digital twin from the physical twin when AI simulation takes place.

When it comes to \textit{updating and maintaining} the simulators, we find the patterns reported in \tabref{tab:patterns-sim}.
\xofy{11}{22}{} sampled approaches implement a continuous, automated update mechanism. As shown in the corresponding architecture in \figref{fig:patterns-sim-auto}, an automated update mechanism implements the \textbf{C-D} loop using \textit{Passive} observation, to which the response is voluminous \textit{Low context} data which the \textit{Digital Twin} has to sift through before \textit{Updating} the model.

This mechanism is contrasted with on-demand techniques that account for \xofy{2}{22}{} cases in our sample.
As shown in \figref{fig:patterns-sim}, on-demand mechanisms respond to situations in which the digital twin cannot provide sufficient simulated data, e.g., due to the \textit{Query} of the \textit{AI} being outside the validity range of the simulator. For example, in \cite{vila2023design}, the reinforcement learning agent asks for the simulation of a state that the simulator has limited or no data about. In these situations, the \textit{Digital Twin} needs to sample from the \textit{Physical Twin}, either in an asynchronous (\figref{fig:patterns-sim-ond-async}) or synchronous fashion (\figref{fig:patterns-sim-ond-sync}). In both cases, \textit{identifying missing data} is the first step, from which an \textit{Active} implementation of the \textbf{C-D} loop follows. \textit{Active} observation is achieved by \textit{Controlling} the \textit{Physical Twin} appropriately.
In response, the observation provides \textit{High context} data, which is more related to the particular action the \textit{Digital Twin} has taken to sample the behavior of its physical surroundings.
As customary in synchronous modes of operation, the execution of AI training might be \textit{Blocked} until the update is complete.

\phantom{}

\begin{conclusionframe}{RQ4: Simulator lifecycle models}
Only about 60\% of digital twin-driven AI simulation techniques support the \textit{maintenance} of the simulator's quality and fidelity. Most techniques implement \textit{automated, passive data collection} from the physical twin for this purpose.
\end{conclusionframe}

\subsection{Technologies and techniques in support of DT-enabled AI simulation (RQ5)} \label{sec:rq6}

We now report on the technological enablers of AI simulation by DTs by first looking at the typical technological choices (\secref{sec:tech-enablers}), and then, investigating technologies and techniques to address challenges specific to digital twins (\secref{sec:challenges-dt}) and AI (\secref{sec:challenges-ai}) components.

\subsubsection{Technological choices at a glance} \label{sec:tech-enablers}

\begin{table*}[t]


\centering
\caption{Software technology: Programming Languages}
\label{tab:software-tech-languages}
\begin{tabular}{@{}p{4cm}lp{11cm}@{}}
\toprule
\multicolumn{1}{c}{\textbf{Languages}} & \multicolumn{1}{c}{\textbf{\#Studies}} & \multicolumn{1}{c}{\textbf{Studies}} \\ \midrule

Python & \maindatabar{12} &
\cite{alexopoulos2020digital,david2023digital,guo2023intelligent,hammar2023digital, li2022when, matulis2021robot,pun2023neural,shen2022deep, shui2023cell,tang2023digital,xia2021digital,zhang2024automated} \\

C\# & \maindatabar{5} & \cite{li2022when,matulis2021robot,xia2021digital,yang2024social,zhang2024automated} \\

Matlab & \maindatabar{3} & \cite{david2023digital,liu2022digital,tubeuf2023increasing} \\

Javascript & \maindatabar{1} & \cite{hammar2023digital} \\

Bash & \maindatabar{1} & \cite{hammar2023digital} \\

R+ (RoboPlus) & \maindatabar{1} & \cite{verner2018robot} \\


Unclear & \maindatabar{6} & \cite{cui2023digital,deng2021digital,dong2019deep,li2023optimization,vila2023design, zhang2023digital} \\
\bottomrule
\end{tabular}


\end{table*}
\begin{table*}[t]


\centering
\caption{Software technology: Frameworks}
\label{tab:software-tech-frameworks}
\begin{tabular}{@{}p{4cm}lp{11cm}@{}}
\toprule
\multicolumn{1}{c}{\textbf{Frameworks}} & \multicolumn{1}{c}{\textbf{\#Studies}} & \multicolumn{1}{c}{\textbf{Studies}} \\ \midrule

TensorFlow & \maindatabar{6} & \cite{alexopoulos2020digital,guo2023intelligent,matulis2021robot,shui2023cell,tang2023digital,zhang2024automated} \\

PyTorch & \maindatabar{4} & \cite{li2022when,pun2023neural,shen2022deep,zhang2024automated} \\

DL Toolbox & \maindatabar{1} & \cite{liu2022digital} \\

RL Toolbox & \maindatabar{1} & \cite{tubeuf2023increasing} \\



Unclear & \maindatabar{12} & \cite{cui2023digital,david2023digital,deng2021digital,dong2019deep,hammar2023digital,li2022when,li2023optimization,verner2018robot,vila2023design,xia2021digital,yang2024social,zhang2023digital} \\
\bottomrule
\end{tabular}


\end{table*}
\begin{table*}[t]


\centering
\caption{Hardware technology}
\label{tab:hardware-tech}
\begin{tabular}{@{}p{4cm}lp{11cm}@{}}
\toprule
\multicolumn{1}{c}{\textbf{Hardware}} & \multicolumn{1}{c}{\textbf{\#Studies}} & \multicolumn{1}{c}{\textbf{Studies}} \\ \midrule

GPU & \maindatabar{7} & \cite{guo2023intelligent,li2022when,pun2023neural,shen2022deep,tang2023digital,yang2024social,zhang2024automated} \\

CPU & \maindatabar{4} & \cite{guo2023intelligent,liu2022digital,yang2024social,zhang2024automated} \\

IoT & \maindatabar{3} & \cite{alexopoulos2020digital,guo2023intelligent,verner2018robot} \\

Others & \maindatabar{6} & \cite{cui2023digital,matulis2021robot,pun2023neural,verner2018robot,xia2021digital,yang2024social} \\

Unclear & \maindatabar{9} & \cite{david2023digital,deng2021digital,dong2019deep,hammar2023digital,li2022when,shui2023cell,tubeuf2023increasing,vila2023design,zhang2023digital} \\
\bottomrule
\end{tabular}


\end{table*}

Tables \ref{tab:software-tech-languages}--\ref{tab:software-tech-frameworks} summarize the typical \textbf{software} choices in the sampled studies.
\tabref{tab:software-tech-languages} reports on the employed programming languages. The majority of the sampled studies (\xofyp{12}{22}) choose Python as their preferred programming language. Some moderately encountered programming languages are C\# and Matlab, which we find in \xofy{5}{22}{} and \xofy{3}{22}{} studies, respectively.
\tabref{tab:software-tech-frameworks} reports on the frameworks we find the sampled studies. We observe that Python-based ML/AI frameworks are prominent, especially TensorFlow\footnote{\label{fn:tensorflow}\url{https://www.tensorflow.org/}} and PyTorch\footnote{\label{fn:pytorch}\url{https://pytorch.org/}}. TensorFlow is slightly more popular than PyTorch, found in \xofy{6}{22}{} and \xofy{4}{22}{} studies, respectively. These open-source, general-purpose frameworks are used for a variety of ML/AI tasks, particularly for training AI models.
Additionally, we find proprietary frameworks from MathWorks, specifically their Deep Learning Toolbox\footnote{\label{fn:dltoolbox}\url{https://www.mathworks.com/products/deep-learning.html}} and Reinforcement Learning Toolbox.\footnote{\url{https://www.mathworks.com/products/reinforcement-learning.html}} \textcite{liu2022digital} use the former to create and train neural networks; and \textcite{tubeuf2023increasing} use the latter framework's SARSA agent to control blowout processes.

\tabref{tab:hardware-tech} summarizes the typical \textbf{hardware} choices in the sampled studies.
We observe that the majority of studies employ traditional hardware elements for developing AI model, such as GPUs (\xofyp{7}{22}) and CPUs (\xofyp{4}{22}). GPUs are well-suited to parallel processing, allowing for faster execution of large-scale matrix calculations required for neural network training, whereas CPUs are less optimized for this type of activity. Consequently, deep learning-based techniques, such as deep reinforcement learning~\cite{guo2023intelligent}  and deep neural network~\cite{pun2023neural}, heavily rely on GPUs.
Internet of Things (IoT) is mentioned only in \xofy{3}{22} studies. For instance, \textcite{verner2018robot} use the IoT platform ThingWorx\footnote{\url{https://www.ptc.com/en/products/thingworx}} to link a robot to the remote computer (i.e., the digital twin), collect and analyze data from a simulator, and provide actuation recommendations to the robot.
Other, sporadically encountered hardware technologies include LiDAR---e.g., by \textcite{pun2023neural}, who use a moving platform with LiDAR, cameras, and GPS to build a more realistic, scalable, and diverse simulation environment---and edge servers---e.g., by \textcite{cui2023digital}, to pre-process sampled physical data on end-points of wireless networks before they get propagated to a digital twin.

\begin{table*}[t]

\centering
\caption{Technologies and techniques for solving DT-specific challenges}
\label{tab:dtchallenges-solutions}
\begin{tabular}{@{}p{2.75cm}p{4.5cm}p{6.25cm}p{2.5cm}@{}}
\toprule
\multicolumn{1}{c}{\textbf{Challenge Group}} & \multicolumn{1}{c}{\textbf{Specific Challenge}} & \multicolumn{1}{c}{\textbf{Technology/Technique}} & \multicolumn{1}{c}{\textbf{Primary Studies}} \\ 
\midrule

\textbf{Fidelity} & Low fidelity  & Calibration & \cite{tang2023digital, verner2018robot} \\
&  & Siemens Process Simulate & \cite{xia2021digital} \\ \cmidrule{2-4}
& Low accuracy in DT modeling & Simplified rendering methods: Diffuse-only reconstruction, separate lighting prediction, and fixed base materials & \cite{alexopoulos2020digital, pun2023neural} \\ \cmidrule{2-4}
& Real-world spatial distributions issues of mobile devices & DBSCAN for high-density identification \& K-means for UAV deployment optimization & \cite{guo2023intelligent} \\ \midrule

\makecell[tl]{\textbf{Latency}\\\textbf{\& Delay}} & Delays in network state updates & Network state prediction via time-series algorithms & \cite{deng2021digital} \\ \cmidrule{2-4}
& Delays in real CSI data collection & Time-series data generation& \cite{shui2023cell} \\ \cmidrule{2-4}
& Simulation latency & Upgraded computing hardware \& lower simulation resolution & \cite{xia2021digital} \\ \cmidrule{2-4}
& Communication latency in networked DT systems & Ethernet connections with iPerf-based latency analysis & \cite{yang2024social} \\ \midrule

\textbf{Limited Data} & Insufficient CSI data for beam search optimization & Data-driven DT with deep learning to map partial RSRPs to full RSRPs & \cite{li2023optimization} \\ \midrule

\makecell[tl]{\textbf{Bandwidth}\\ \textbf{Limitations}} & Real-time DT updates impacted by bandwidth constraints & Distributed nodes \& MQTT pub-sub mechanism \& Prioritized communication channels & \cite{alexopoulos2020digital} \\ 
\midrule

\textbf{Data Security} & Prevent data leakage and message tampering & Blockchain \& data consistency authentication & \cite{liu2022digital} \\

\bottomrule
\end{tabular}

\end{table*}

\subsubsection{Technologies and techniques for addressing digital twin-specific challenges}\label{sec:challenges-dt}

\tabref{tab:dtchallenges-solutions} summarizes how different technologies and techniques contribute to tackling challenges specific to digital twins. 
Such challenges primarily include ensuring sufficient fidelity (\xofyp{6}{22}{}) and lower latency of communication between physical and digital twins (\xofyp{4}{22}{}). Both concerns impact the accuracy, reliability, and efficiency of digital twin models.

To improve fidelity, virtual models need to be calibrated. \textcite{tang2023digital} implement real-time calibration of action and state spaces based on transmission delay analysis.
\textcite{verner2018robot} employ center of gravity and sensitivity analysis in Creo\footnote{\url{https://creo.ca/en/}} to calibrate the model.
\textcite{xia2021digital} use Siemens Process Simulate\footnote{\url{https://plm.sw.siemens.com/en-US/tecnomatix/process-simulate-software}} to ensure high fidelity.
Sometimes, fidelity boils down to topological details of physical twins. For example, \textcite{guo2023intelligent} face fidelity issues that stem from the real-world spatial distributions of mobile devices in a network area; and tackle the fidelity issues by various machine learning techniques, such as density-based spatial clustering of applications with noise (DBSCAN) and k-means clustering, using the scikit-learn library.\footnote{\url{https://scikit-learn.org/stable/}}

Latency concerns in digital twin-based AI simulation result from delays in state changes, communication, and simulation execution.
Time series-based approaches can help reduce latency by predicting and anticipating system states. Such methods are particularly popular in the networks domain. For example, \textcite{deng2021digital} employ time-series algorithms for network state prediction, and \textcite{shui2023cell} use time-series data generation to overcome delays in real channel state information (CSI) data collection. Another notable form of latency manifests in simulation. Digital twin-based techniques are often meant to provide rapid decision support and control, and delays in obtaining simulation results might be costly. For example, \textcite{xia2021digital} handle simulation latency by updating computing hardware and lowering simulation resolution.

Apart from fidelity and latency, some additional digital twin-specific challenges include limited data availability, bandwidth limitations, and security.
To handle limited data availability, \textcite{li2023optimization} use deep learning-based data-driven digital twins to infer missing channel state information (CSI) data.
To overcome bandwidth limitations in real-time digital twin updates, \textcite{alexopoulos2020digital} employ an array of techniques, including distributing nodes for decentralized data transmission, MQTT’s pub-sub mechanism for efficient data exchange, and prioritized communication channels to optimize network traffic flow. To ensure data security, e.g., by mitigating data leakage and message tampering risks, \textcite{liu2022digital} use blockchain and data consistency authentication techniques to ensure data integrity and traceability.

\begin{table*}[t]

\centering
\caption{Technologies and techniques for solving AI-specific challenges}
\label{tab:aichallenges-solutions}
\begin{tabular}
{@{}p{2.75cm}p{4.5cm}p{6cm}p{2.75cm}@{}}
\toprule
\multicolumn{1}{c}{\textbf{Challenge Group}} & \multicolumn{1}{c}{\textbf{Specific Challenge}} & \multicolumn{1}{c}{\textbf{Technology/Technique}} & \multicolumn{1}{c}{\textbf{Primary Studies}} \\ 
\midrule

\makecell[tl]{\textbf{AI Model}\\ \textbf{Optimization}} & Difficult to get the state transition probability for RL & Model-free RLs & \cite{li2022when, liu2022digital, matulis2021robot, shen2022deep, zhang2023digital} \\ \cmidrule{2-4}
& Over-exploitation & User actuation to prevent undeserved AI rewards & \cite{matulis2021robot} \\
&  & Intelligent Social Learning (ISL) integrated with SAC & \cite{yang2024social} \\ \midrule

\makecell[tl]{\textbf{AI Model}\\ \textbf{Convergence}} & High-dimensional search space (Curse of Dimensionality) & Parallel edge-trained agents& \cite{cui2023digital} \\ \cmidrule{2-4}
& High-dimensional action space & Multi-agent tasks for decentralized control & \cite{li2023optimization} \\ \cmidrule{2-4}
& Slow Convergence & Experience Replay Buffer & \cite{li2022when, shen2022deep, tang2023digital,  zhang2024automated}\\ 
\midrule

\makecell[tl]{\textbf{High}\\ \textbf{Comp. Load}} & High power and time overhead & Decomposing the optimization model into two subproblems & \cite{liu2022digital} \\ \cmidrule{2-4}
& High computational load due to limited processing power & Task decomposition using a genetic algorithm, divides a task into multiple subtasks & \cite{tang2023digital} \\
\midrule

\makecell[tl]{\textbf{AI Prediction}\\ \textbf{Accuracy}} & Difficulty in predicting UAV trajectories & LSTM-based future state estimation & \cite{li2022when, shen2022deep} \\ \midrule

\makecell[tl]{\textbf{AI System}\\ \textbf{Robustness}} & Prevent congestion and deadlock in multi-agent systems & A* static path planning + DRL-based dynamic collision avoidance & \cite{zhang2024automated} \\ 
\bottomrule
\end{tabular}

\end{table*}

\subsubsection{Technologies and techniques for addressing AI-specific challenges}\label{sec:challenges-ai}

\tabref{tab:aichallenges-solutions} summarizes how different technologies and techniques contribute to tackling key AI-specific challenges.
Such challenges primarily include AI model optimization (\xofyp{7}{22}{}) and AI model stability (\xofyp{6}{22}{}).

Optimizing AI algorithms is a fundamental challenge in AI simulation. In RL, optimizing policy learning is challenging when state transition probabilities are unknown in real-world cases. To overcome this, model-free RL techniques, such as Double Deep Q-Network (DDQN)~\cite{liu2022digital}, Proximal policy optimization (PPO)~\cite{matulis2021robot}, Deep Deterministic Policy Gradient (DDPG)~\cite{li2022when, shen2022deep}, and Soft Actor Critic (SAC)~\cite{zhang2023digital}, are used to optimize learning by directly interacting with the environment without requiring explicit transition probabilities. We find various frameworks in our sample to implement these techniques, including MathWorks' Deep Learning Toolbox for DDQN, TensorFlow for PPO, and PyTorch for DDPG. Over-exploitation is another challenge in RL-based model optimization, as it causes agents to over-rely on past experiences, leading to suboptimal policies that fail to generalize effectively. \textcite{matulis2021robot} tackle this by using user actuation, where human observers monitor and adjust AI behavior in virtual environments to prevent agents from exploiting unintended flaws or loopholes for rewards. This ensures learned behaviors remain realistic and generalizable to real-world scenarios. \textcite{yang2024social} use GPU-accelerated Intelligent Social Learning (ISL), a learning combination of three strategies: guided learning from top-performing agents, imitation with controlled randomness, and independent self-exploration. This structured learning approach helps agents escape local optima and improve decision-making in sparse and noisy environments.

Ensuring convergence, i.e., the efficient progression of models toward an optimal solution is another key challenge in AI that pertains to AI simulation. One key factor that hinders convergence is high-dimensional search spaces, which slow learning and convergence by exponentially increasing possibilities. \textcite{cui2023digital} address this by deploying parallel agents on an edge server, where each agent is trained closer to real-time data sources and focuses on a smaller subset of the problem. This distributed approach accelerates convergence by enabling agents to learn in parallel and interactively refine their policies. Beyond high-dimensional search spaces, high-dimensional action spaces further hinder convergence by exponentially increasing the number of possible actions an agent can take at each step. As decision complexity grows, models struggle to explore and update policies effectively, leading to slower convergence and potential divergence. \textcite{li2023optimization} tackle this by using multi-agent tasks, where multiple agents control different parts of the action space instead of a single centralized agent. By distributing decision-making among multiple agents, this approach reduces the burden on any single model, allowing for steadier policy updates and smoother convergence of the overall system. In addition to addressing high-dimensional learning challenges, \xofyp{4}{22}{} studies tackle slow convergence and inefficient learning problem by using experience replay buffer (i.e., memory mechanism for storing and reusing past interactions). \textcite{li2022when} and \textcite{zhang2024automated} use the buffer to speed up convergence by allowing reinforcement learning models to store and reuse historical experiences, leading to more effective learning with fewer training samples.  \textcite{shen2022deep} store the training data in the buffer and randomly sample from it, ensuring the data remains independent and identically distributed (IID), which prevents overfitting to sequential dependencies and accelerates convergence. \textcite{tang2023digital} store multi-agents' interactions, joint actions and states in a common buffer, leading to more synchronized training and faster convergence. These four studies not only illustrate the role of experience replay in improving convergence but also highlight the importance of Python and GPU acceleration in efficiently training models. Among them, \textcite{tang2023digital} use TensorFlow, while the others use PyTorch.

Apart from optimization and stability, some additional AI-specific challenges include the increased computational load of AI, prediction accuracy, and system robustness. To reduce computational overhead, \textcite{liu2022digital} and \textcite{tang2023digital} use decomposition techniques to break down tasks. \textcite{liu2022digital} decompose the model with two subproblems handled by decision tree algorithms (DTA) and double deep Q-learning (DDQN). \textcite{tang2023digital} use genetic algorithm-based task decomposition, dividing tasks into subtasks that can be processed in parallel, reducing overall computational overhead. AI prediction accuracy is challenging in Unmanned Aerial Vehicle (UAV) systems due to unpredictable motion dynamics. \textcite{li2022when} and \textcite{shen2022deep} address this by implementing long short-term memory (LSTM) networks, implemented in Python, to capture temporal dependencies in sequential data. This enables the model to learn patterns in UAV trajectories and improve future state predictions. System robustness (i.e., an AI system’s ability to function reliably under uncertainties and dynamic conditions) is another critical concern. \textcite{zhang2024automated} integrate a static path planning agent using A* algorithm to compute globally optimized, conflict-free paths and a DRL-based dynamic collision avoidance agent to adaptively adjust to obstacles in real-time, enhancing the overall robustness of the system. Both agents are implemented in PyTorch and trained using a CPU.

\phantom{}

\begin{conclusionframe}{RQ5: Technologies and techniques}
AI simulation predominantly uses \textit{Python}, with the \textit{TensorFlow} being the most popular, and commonly relying on \textit{GPU} processing. 
Digital twin-related challenges are tackled through calibration for improving fidelity, time-series prediction for mitigating latency, and blockchain for enhancing security. AI-related challenges are tackled through model-free RL for optimizing AI models, experience replay buffer for improving model convergence, and decomposition-based task allocation for high computational load.
\end{conclusionframe}

\subsection{Open Challenges (RQ6)}\label{sec:rq5}

We now discuss key challenges mentioned in the primary studies.
We warn that challenges and limitations are sporadically reported (see \figref{fig:pubtrends}). To mitigate threats to validity, we avoid interpretation at this point as much as possible and report only factual information.

\subsubsection{Fidelity and other extra-functional properties}

Obviously, fidelity is a key property of simulated data, directly linked to the fidelity and accuracy of the digital twin's simulation model~\cite{alexopoulos2020digital}; but \textbf{fidelity is hard to assess and ensure}~\cite{shui2023cell}.
\textcite{shen2022deep} highlight the challenges involved in achieving accurate virtual replications. Sim-to-real transfer is particularly challenging, as noted by \textcite{li2022when} who state that ``\textit{the gap between simulation and reality greatly limits the application of deep reinforcement learning in the path planning problem of multi-UAV}.'' \textcite{pun2023neural} recognize sim-to-real discrepancies, particularly when encoding simulation models in generative adversarial networks (GANs).
Among other extra-functional properties, \textbf{safety}~\cite{zhang2024automated}, \textbf{reliability}~\cite{dong2019deep}, and \textbf{security}~\cite{tubeuf2023increasing} are mentioned. For example, reliability is a particular concern in multi-access edge computing (also known as mobile edge computing)~\cite{dong2019deep} due to its ultra-low latency guarantees.

\subsubsection{Interactions with the physical system}

\textcite{shui2023cell} warn that the \textbf{frequency of interactions} between the digital and physical twin might be limited and thus, inadequate for acquiring sufficient amounts of real data.
Similar problems have been voiced by \textcite{shen2022deep}.
In some cases, data might be provided by human stakeholders~\cite{david2023digital}, naturally limiting the update frequency of the model and the quality of collected data.

\subsubsection{Process aspects}
In general, the \textbf{transition from concept to practical implementation} of digital twins is recognized as a complex process.
\textcite{matulis2021robot} voice concerns over the complexity of real-life manufacturing settings which challenges deployment. \textcite{tubeuf2023increasing} mention the challenges of deploying overly sophisticated models into real settings.
Among the frequent organizational challenges, implementation expenses and organizational maturity levels are cited. \textcite{alexopoulos2020digital} identify development and integration as the key cost factors in their DT-enabled AI simulation approach. \textcite{david2023digital} report mismatches between twinning ambitions and low levels of operational maturity from underdigitalized domains, such as cyber-biophysical systems.

\subsubsection{Challenges as boundary conditions} There are challenges that are outside of the expertise of DT and MDE experts. These challenges are to be treated as boundary conditions in prospective projects.
Multiple studies cite the \textbf{elevated computational and hardware demands} of DT-enabled AI simulation. Hardware constraints and inadequate computing power substantially impacts AI training, as noted by \textcite{matulis2021robot}. Storage space might be a limitation as well, especially in solutions running through external cloud-hosted services, as discussed, e.g., by \textcite{deng2021digital}.
Multiple studies mention the \textbf{tuning challenges of AI algorithms}, with typical examples of finding the trade-off between exploration and exploitation in reinforcement learning~\cite{xia2021digital} and fine-tuning the hyperparameters of deep learning~\cite{tubeuf2023increasing}.

\phantom{}

\begin{conclusionframe}{RQ6: Open challenges}
Challenges in DT-enabled AI simulation include both \textit{technical} (e.g., assessing and ensuring fidelity and establishing sufficient interactions with the physical twin) and \textit{organizational} kinds (e.g., managing development processes).
\end{conclusionframe}
\section{Discussion}\label{sec:discussion}

We now discuss the key takeaways and lessons learned.

\subsection{Key takeaways}

\subsubsection{Digital twinning brings unique benefits (and challenges) to AI simulation}

Digital twinning seems to be a useful instrument in implementing AI simulation. As a key benefit, digital twins provide mature system organization principles and architectures in which the key components of AI simulation be situated---simulators as clearly defined functional entities~\cite{shao2021use} and AI as a service through well-defined end-points~\cite{david2023digital}. Another benefit of digital twins is their uniquely tight coupling with the underlying physical systems, which digital twins can observe (\figref{fig:patterns-sim-auto} -- e.g., \cite{cui2023digital,deng2021digital}) and interrogate upon request (Fig. \ref{fig:patterns-sim-ond-sync}--\ref{fig:patterns-sim-ond-async} -- e.g., \cite{alexopoulos2020digital,vila2023design}), allowing for evolutionary strategies of simulators.
As evidenced by \tabref{tab:activities-ai}, the majority of the sampled works (\xofyp{14}{22}{}) go beyond simple data generation and labeling, and develop virtual training environments, in which evolutionary maintenance of simulation models is a typical feature.
On the negative side (\secref{sec:rq5}), fidelity and proper lifecycle models for digital twins remain a challenge.

These demonstrated contributions to the surging AI market suggest a likely increased adoption rate of digital twin technology. We anticipate digitally adept domains to follow suit with networking and robotics (\secref{sec:rq1}) and adopt digital twins for AI simulation and traditional control and governance-related purposes. Thus, the link between digital twins and AI is shaping up to be one of the impactful directions for digital twin researchers.

\subsubsection{(Wireless) networks and robotics paving the way for DT-enabled AI simulation}

There is a clear trend in the application domains of DT-enabled AI simulation, with networks and robotics accounting for \xofy{17}{22}{} of the sampled approaches (see \secref{sec:rq1}). 
These numbers are rather unexpected after the recent cross-domain systematic mapping study on software engineering for digital twins by \textcite{dalibor2022cross}, who do not mention these fields as frequent adopters of digital twins~\cite[Fig. 5]{dalibor2022cross}. Granted, robotics might fit the ``manufacturing'' in that classification, but the emergence of the networking domain as a top adopter suggests a shift in tone-setters as DT-enabled AI simulation might be growing out of domains different from traditional digital twinning.
The strong showing of robotics might be explained by digital twinning being an already adopted technology. The high research activity in networking seems to be a transformative tendency, potentially due to the relative lack of digital twinning impediments~\cite[Sec. 3.3]{trauer2022challenges} in the domain.

\subsubsection{Genuine digital twins dominate AI simulation}

One of the unexpected observations of this study is the strong alignment of the notion of a digital twin with the classical definitions~\textcite{kritzinger2018digital}. \xofy{19}{22}{} studies (\secref{sec:rq2}) report a digital twin that (i) collects real-time from a physical system and (ii) exerts control on the physical system. This number is much higher than in traditionally considered digital twinning domains, such as manufacturing, where ``digital shadows'' are quite often encountered.
We hypothesize that the recent surge (2022--2024) of digital twinning in the network domain benefited from mature technologies in an already highly digitalized domain, allowing for advanced digital twin solutions.

\subsubsection{Deep learning proliferates -- and that, in different flavors}

The main observation in regard to RQ3 (\secref{sec:rq3}) is that reinforcement learning is particularly highly utilized (e.g., \cite{verner2018robot,yang2024social}). We see reinforcement learning as a naturally good fit with twinned setups. Reinforcement learning relies on a trial-and-error learning Markovian process~\cite{sutton2018reinforcement}, in which digital twins can act as the supporting technology for safe, reliable, and reproducible experiments. This role is in line with the ``risk-free experimentation aid'' role of DTs envisioned by \textcite{barat2022digital} in techno-socio-economic systems.

Within reinforcement learning, we find a high number of \textit{deep} reinforcement learning methods (e.g., \cite{guo2023intelligent,li2022when}), that is, reinforcement learning that encodes the policy as a deep neural network. This number, \xofy{13}{22}{}, together with other deep learning techniques (e.g., \cite{dong2019deep,pun2023neural}) amounting to \xofy{4}{22}{} studies, means that a total of \xofy{17}{22}{} methods rely on deep neural networks. Thus, DTs have to be able to provide large amounts of data, either in small batches through rapid interactions (\figref{fig:patterns-ai-rl}) or as big data at once (\figref{fig:patterns-ai-dl}). Both scenarios challenge extra-functional quality metrics of DT, such as performance, reliability, and availability~\cite{tubeuf2023increasing}.

\subsubsection{Simulation: ``using the most appropriate formalisms''}

The choice of modeling and simulation formalisms aligns with the distribution of domains (\secref{sec:rq1}). We see a number of network models describing topologies and channel dynamics (e.g., \cite{liu2022digital,shui2023cell}), used in network-themed studies. We see a number of physics and CAD 3D geometry models in robotics and manufacturing-themed studies (e.g., \cite{shen2022deep,verner2018robot}), which is in line with the observations of \citeauthor{dalibor2022cross}, who report a high number of CAD 3D models and mathematical physical models in their systematic mapping study~\cite[Fig. 11]{dalibor2022cross}.

In some cases, however, the exact simulation formalism is hard to identify. These are the cases in which the simulation model itself is encoded in a neural network, such as a deep neural network (e.g., \cite{li2023optimization}) or a generative adversarial network (GAN) (e.g. \cite{pun2023neural}).

\subsubsection{Technology: reliance on traditional choices}\label{sec:discussion-techchoices}

The data disclosed in \secref{sec:rq6} demonstrates a particularly strong reliance on traditional technology, especially in terms of AI. Most modern hardware and software solutions in support of computationally demanding AI are lacking. Following the taxonomy of hardware accelerators by \textcite{peccerillo2022survey}, hardware in the sampled studies (\tabref{tab:hardware-tech}) is exclusively general-purpose, i.e., features components (processors, memory, etc.) that can be normally found in a general-purpose computing system. Notable lacking components are, for example,
graphics-oriented DRAM technologies (e.g., graphics double data rate 6 SDRAMs),
high speed, 3D-stacked DRAM memory technologies (e.g., multi-channel DRAMs, and hybrid memory cube DRAMs),
and specialized application-specific integrated circuits (ASIC), such as Tensor Processing Units (TPUs). This reveals a significant deficiency of contemporary DT-enabled AI simulation in terms of AI-specialized technology. This, in turn, foreshadows potential limitations of AI simulation when state-of-the-art AI agents are to be trained by current AI simulation methods. Particularly, the increased computing power of properly-built AI agents might put additional pressure of the already demonstrated challenges of computational and hardware demands of simulators (\secref{sec:rq5}).

In AI/ML frameworks, we TensorFlow being heavily favored over PyTorch (\tabref{tab:software-tech-frameworks}), which is the opposite of the trends observed in AI/ML research~\cite{pytorch-tensorflow-2023}. This observation can be the artifact of long-running projects our sampled studies report on, with some outdated technology; or the artifact of the maturity of solutions the sampled studies report on. PyTorch is more often used in research due to its flexibility and ease of use, while TensorFlow is often used for production applications due to its speed and scalability~\cite{pytorch-tensorflow-2023}.

In terms of DTs, we see a complete lack of DT platforms and frameworks. Both popular open-source (e.g., Eclipse Ditto and BaSyx) and proprietary platforms (e.g., Microsoft's Azure Digital Twins and Amazon's AWS IoT TwinMaker) are missing. We hypothesize that in domains from which our corpus is sampled from, either lack the expertise in configuring and maintaining extensive DT platforms, or the benefits of DT platforms are not well-understood.

\subsection{Lessons learned for the DT and MDE Communities}

\subsubsection{Architectural concerns}
Perhaps the most important lesson learned for the DT and MDE communities is the complete lack of digital twin standards, architectural blueprints, and reporting guidelines in the primary studies we sampled. As reported in \secref{sec:rq2}, we found only one study (4.5\%) that relies on the Reference Architectural Model Industrie 4.0 (RAMI4.0)~\cite{rami-website}, but even in this sole case~\cite{alexopoulos2020digital}, the work failed to make a connection with the Asset Administration Shell (AAS)~\cite{aas-website}, the standardized digital representation of assets within RAMI for digital twinning purposes.
The lack of architectural standardization is particularly concerning in cases when legacy systems are retrofitted to accommodate digital twins, and the ramifications of system evolution are not being investigated at the architectural level.
Standards, such as the ISO 23247 Digital Twin Framework for Manufacturing~\cite{shao2021use}, hold particular potential in this aspect and should be considered by prospective researchers.
We recommend the DT community to focus efforts on \textbf{making DT architectural standards more accessible to AI researchers} for the sake of scalable, reliable, and sustainable AI simulation.

\subsubsection{Towards better technical sustainability of AI simulation by digital twins}

Technical sustainability is the ability of a system to be used over an extended lifetime~\cite{penzenstadler2013generic}. In terms of AI simulation, prolonged usability boils down primarily to maintaining the faithfulness and validity of simulators. The general notion of AI simulation does not consider this longitudinal dimension~\cite{aisim-gartner}. Digital twins improve the technical sustainability outlooks of AI simulation by construction. It is thanks to the tight coupling with its physical environment that digital twins can support various modes of maintaining their simulators' faithfulness, e.g., through observing or experimenting with the physical environment (\secref{sec:rq4}). Recent developments in digital twin evolution
\cite{david2023towards,michael2024smart}
provide additional support for the technical sustainability of AI simulation.

In this respect, we note a low number of techniques that implement on-demand simulator maintenance (\tabref{tab:patterns-sim}), as only about 18\% of the sampled studies do so. We \textbf{recommend researching more sophisticated maintenance mechanisms, architectures, and lifecycle models} (Fig. \ref{fig:patterns-sim-ond-async}--\ref{fig:patterns-sim-ond-sync}) in response to the anticipated demand for such features. We warn that these efforts might be challenged by the lack of standards which we observed in \secref{sec:rq2}, and which is an acute issue in digital twin engineering in general~\cite[Sec 6.3.3]{muctadir2024current}.

\subsubsection{Validity and sim2real}

Increasing efforts have been dedicated to transferring the knowledge obtained in a simulated environment to real-world applications, known as sim-to-real~\cite{doersch2019sim2real}---a potential problem in critical systems, such as autonomous vehicles~\cite{hu2024how}.
Our investigation of the state of the art reveals that little attention is dedicated to the sim-to-real problem in digital twin-based AI simulation currently. (See the replication package for more details.)

The validity of models has been of particular interest in the modeling and simulation community. Especially in recent years, the traditional and vague notion of a simulation frame by \textcite{zeigler2018theory} has been clarified by a series of works.  \textcite{biglari2022model}, \textcite{mittal2023towards}, and \textcite{acker2024validity} situate validity at the digital-to-physical boundary of digital twins, reflecting on the validity of simulation models w.r.t. to environmental conditions, engineering assumptions, etc. We recommend \textbf{modeling and simulation experts to adopt research results on validity frames in support of AI simulation} to allow for better sim-to-real transfer. 

\subsubsection{Human factors in AI simulation \label{humanfactors}} We observe an overall ignorance of human factors. This holds both for human experts in the AI simulation loop and for human stakeholders in the development and operation of digital twins serving AI simulation. These trends are best exemplified in \secref{sec:rq2} and, specifically, in the breakdown of system organization patterns in \tabref{tab:patterns-arch}. These trends are not entirely surprising as socio-technical views on digital twins are in their early phase~\cite{david2024infonomics}. The role of the human in the loop is fully expected to grow~\cite{shajari2024twin}, e.g., in training the virtual replicas~\cite[Fig 7]{muctadir2024current}, and guiding AI agents in their learning phase~\cite{dagenais2024opinion-techreport}. With that, we recommend digital twin experts to \textbf{adopt more human-centered views on digital twins, both in terms of the human as an interactive user of AI simulation and as a stakeholder in the development and operation process of DTs}.

\subsubsection{Reporting quality and recommendations}
Finally, we remark on some quality-related trends in the primary studies in our corpus. First of all, we notice a low level of detail in discussing the simulation aspects of AI simulation (\secref{sec:pub-trends}). This is a severe shortcoming, considering the central role of simulation in these approaches. The lack of detail about simulation is especially concerning, given that the validity of simulation models is the primary factor that determines the validity of data that is generated for training AI agents. We recommend prospective researchers to \textbf{be more detailed and transparent about simulation formalisms, methods, algorithms, and tools} when reporting their work. This will allow for independent validation and reproduction of results.

We also note the staggering lack of support for the reproducibility and independent validation of results. We have not found any data supplements or replication packages despite data being the central artifact in AI simulation. We urge \textbf{methodologists in simulation and AI to develop joint standards}, and \textbf{conference organizers to introduce artifact evaluation practices}, such as the ACM Artifact Review and Badging procedure~\cite{acm-badge}.
\begin{table*}[]

\centering
\scriptsize
\caption{Mapping of DT4AI concepts onto the ISO 23247 reference architecture.}
\label{tab:iso-mapping}
\begin{tabular}{@{}lcccccccccccccccccc@{}}
\toprule
 & \multicolumn{12}{c}{Digital Twin} & \multicolumn{6}{c}{Device Communication} \\ \cmidrule(lr){2-13}\cmidrule(lr){14-19}
 & \multicolumn{4}{c}{Operation\&Mgmt} & \multicolumn{4}{c}{Application\&Service} & \multicolumn{4}{c}{Res. Acc.\&Interchange} & \multicolumn{3}{c}{Data Collection} & \multicolumn{3}{c}{Device Control}  \\
 \cmidrule(lr){2-5}\cmidrule(lr){6-9}\cmidrule(lr){10-13}\cmidrule(lr){14-16}\cmidrule(lr){17-19}
 & Syn & Pre & DRep & Mai & Sim & AnS & Rep & ApS & IS & AC & P\&P & PI & DC & DP & CoI & Ctr & Act & CtrI \\ \midrule

\textbf{Simulator} & & & & & \yes{} & \partially{} & & & & & & & & & & & & \\
\textit{A: Query} &  &  &  &  & \inputTo{} &  &  &  &  &  &  &  &  & & & & & \\
\textit{B: Sim. data} &  &  &  &  & \outputFrom{} &  &  &  &  &  &  &  &  & & & & & \\ \midrule

\textbf{M} & & & \yes{} & & & & & & & & & & & & & & &\\
\textit{C: Observe} & \partially{} &  &  &  &  &  &  &  &  &  &  &  & & & \partially{} & & & \\
\textit{D: Real data} &  &  &  &  &  & \partially{} &  &  &  &  &  &  & \partially{} & \partially{} & & & & \\
\textit{E: Update} &  &  & \inputTo{} &  &  &  &  &  &  &  &  &  &  & & & & & \\
\textit{F: Control} &  &  &  &  &  &  &  &  &  &  &  &  &  & & & \partially{} & \partially{} &  \partially{} \\ \midrule

\textbf{AccCtrl} &  &  &  &  &  &  &  &  &  & \yes{} &  &  &  & & & & & \\
\textit{G: Direct acc.} &  &  &  &  &  &  &  &  &  & \outputFrom{} &  &  &  & & & & & \\ \bottomrule
\end{tabular}
\caption*{\footnotesize{\yes{}: Strongly related~~\partially{}: Partially related~~\inputTo{}: input to~~\outputFrom{}: output from}\vspace{-1em}}
\caption*{\footnotesize{Syn: Synchronization FE; Pre: Presentation FE; DRep: Digital representation FE; Mai: Maintenance FE; Sim: Simulation FE; AnS: Analytic service FE; Rep: Reporting FE; ApS: Application support FE; IS: Interoperability support FE; AC: Access control FE; P\&P: Plug and play support FE; PI: Peer interface FE; DC: Data collecting FE; DP: Data pre-processing FE; CoI: Collection identification FE; Ctr: Controlling FE; Act: Actuation FE; CtrI: Control identification FE.}}

\end{table*}

\section{Mapping the DT4AI reference framework to the ISO 23247 reference architecture}\label{sec:iso-mapping}

In this section, we provide a mapping of the DT4AI reference framework to the digital twin reference architecture defined in the ISO 23247 standard (``Digital twin framework for manufacturing'')~\cite{shao2021use}.\footnote{\url{https://www.iso.org/standard/78743.html}}
While the DT4AI framework allows for a clean alignment of AI simulation and digital twin concepts, it does not provide architectural guidelines for engineering complex, digital twin-enabled AI simulation software. By mapping the framework onto an ISO-standard reference architecture, we aim to provide researchers and practitioners with clear directives in their digital twin software development and standardization endeavors.

\subsection{Mapping DT4AI Components on ISO 23247 Entities and Functional Entities}

\tabref{tab:iso-mapping} shows how the DT4AI framework maps onto the ISO 23247 reference architecture.
We identify strong ({\small\yes{}}) and partial ({\small\partially{}}) correlations between DT4AI concepts and ISO 23247 architectural elements.

The three main components of the DT4AI framework map on the \textit{Entities} of ISO 23247. Specifically, the \textit{Digital Twin} component maps on the \textit{Digital Twin} and \textit{Device Communication Entities}; the \textit{Physical Twin} component maps on the \textit{Observable Manufacturing Elements (OMEs)} Entity; and the \textit{AI} component maps on the \textit{User Entity}.

In the following, we elaborate on the mapping at the level of functional entities (FE), i.e., units of elementary functionality that perform a group of tasks and functions in digital twins~\cite{shao2021use}.


\subsubsection{Simulation}

The \textit{Simulator} of the DT4AI framework maps directly on the \textit{Simulation Functional Entity (FE)} of the ISO 23247 architecture. According to the standard, the Simulation FE predicts the behavior of observable manufacturing elements (OMEs). In this context, the \textit{A: Query} is the \textit{input} to the Simulation FE and the \textit{B: Simulated data} is its \textit{output}. Querying the simulator and receiving simulated data in response are common traits in every flavor AI simulation (\secref{sec:rq3}).

A partially related architectural element is the \textit{Analytic service FE}, which manages the results of simulations. Depending on the interaction protocol between the Simulator and the AI agent, data might be staged in the Analytic service FE before being passed to the AI agent. This might be particularly useful in batch training scenarios, e.g., in deep learning when larger amounts of data are required to be generated and returned to the AI agent (\figref{fig:patterns-ai-dl}).

\subsubsection{Model and its maintenance}

The simulation model \textit{M} of the DT4AI framework maps onto the \textit{Digital Representation FE} of the ISO 23247 architecture, which, according to the standard, models information from an OME to represent its physical characteristics and status, etc.
The \textit{E: Update} link is the one that maintains the model, and thus, it maps as an \textit{input} to the Digital Representation FE.

Observation of, and data collection from the physical twin are achieved by two links in the DT4AI framework.
The \textit{C: Observe} link is partially related to the \textit{Synchronization FE}, which synchronizes the status of the digital twin with the status of the corresponding OME.
To collect the right data, observation is also driven by the related \textit{Collection identification FE}, which identifies the data needed from OMEs. This typically happens when querying a simulator for simulated data outside of its validity range.
The \textit{D: Real data} link is related to two functional entities. First, the \textit{Data collecting FE}, which is responsible for collecting data from OMEs; and second, the \textit{Data pre-processing FE}, which pre-processes the collected data (e.g., by filtering and aggregation). The \textit{D: Real data} link is partially related to the \textit{Analytic service FE}, which manages the data collected from OMEs. Depending on the interaction protocol between the digital twin and physical twin, data might be staged in the Analytic service FE before being used for updating the Model. For example, when the model needs to be updated with smoothed time series data, the digital twin might collect raw data and the Analytic service FE might apply an ARIMA smoothing~\cite{newbold1983arima} before passing the data to the model.

Finally, in the model and maintenance group of elements, the \textit{F: Control} link serves for controlling the physical twin for purposeful experimentation. This link is related to the three FEs in the Device Control Sub-entity of the ISO 23247 architecture. 
First, the \textit{Control identification FE}, which identifies the OMEs to be controlled. Second, the \textit{Controlling FE}, which sends commands to OME devices in the language understood by that device. And finally, the \textit{Actuation FE}, which actuates OMEs in accordance with the control logic.

\subsubsection{Access control}

The \textit{Access control} element of the DT4AI framework maps clearly on the \textit{Access control FE} of the ISO 23247 architecture, which controls the access of the user entity (here: the AI agent) to the physical twin. In this context, the \textit{G: Direct access} link is the \textit{output} of the Access control FE. The AI agent might decide to interact with the physical twin and use the digital twin as a mere proxy, e.g., for calibrating the AI model in a real physical setting after training on AI-simulated data.

\subsection{Critical reflection}

To critically evaluate the potential limitations of the ISO 23247 reference architecture in accommodating the DT4AI framework,
we synthesize evidence from key studies in the area.

A key limitation in the ISO 23247 reference architecture is its \textbf{lack of explicit support for data storage}. \textcite{ferko2023standardisation} highlight that the reference architecture does not provide a specific functional entity for enhanced data storage, despite acknowledging the need for data exchange via databases or cloud systems. This is limitation might be particularly pressing for the flavors of AI simulation in which large volumes of high-velocity data feed learning algorithms. Advanced analytics and deep-learning FEs are also missing from ISO 23247~\cite{kang2024edge}. AI techniques such as Convolutional Neural Networks (CNNs), Recurrent Neural Networks (RNNs), and reinforcement learning (RL) require significant computational resources (e.g., GPU allocations) and continuous training cycles, which differ markedly from standard analytical or simulation tasks. The current standard, however, does not define a dedicated FE to accommodate these AI-driven functions.

\textcite{shao2024digital} identify some key limitations of the ISO 23247 standard that might hinder AI simulation.
Notably, the standard \textbf{does not address verification, validation, and uncertainty quantification (VVUQ)}, which are fundamental to assessing the credibility of digital twin-enabled AI simulations. Ensuring fidelity and validity is critical for guaranteeing that digital twins accurately reflect real-world processes and provide reliable AI simulation.
Additionally, the standard \textbf{lacks a formal ontology to ensure consistent and machine-processable representations} of digital twin entities. By adopting an ontological framework, developers could establish semantic interoperability, ensuring that digital twin models are both interpretable across different systems and domains, and improving model validity and consistency, a crucial aspect for maintaining trust in AI-based simulations. 
\textcite{shao2024digital} also suggest extensions to the ISO 23247 standard that might be useful for AI simulation. One notable proposal is the ISO 23247-6 Digital Twin Composition supplement of the standard, with the idea of \textbf{facilitating the integration of multiple, interoperable digital twins}. Such an extension might be of high utility in multi-agent digital twin-enabled AI simulation. A demonstrative case for such scenarios is given by \textcite{li2022when}, who propose a task assignment method for multi-UAV (Unmanned Aerial Vehicle) systems in which digital twins of individual UAVs need to exchange real-time data, optimize collective task execution, and dynamically adapt to environmental changes.

Beyond these areas, ISO 23247 also leaves room for improvement in \textbf{lifecycle management}. \textcite{wallner2023digital} observe that the standard addresses data exchange between physical and virtual realms but offers minimal guidance on structuring, versioning, or maintaining the digital twin’s own lifecycle stages. A lifecycle meta-layer could help in reconfiguring flexible manufacturing cells, continuously deploying AI models, and conducting robust change analyses. Furthermore, while the standard refers to human and user entities, it provides limited direction on human-centric considerations, which are an important component of Industry 5.0.  Our findings in Section \ref{humanfactors} show that human stakeholders, both as sources of expert knowledge and as end users, require more explicit modeling to ensure socio-technical alignment, usability, and trust in AI-driven systems.

Finally, \textcite{alcaraz2025digital} draw attention to the potential \textbf{security and privacy concerns} in digital twin technology while evaluating multiple reference architectures, including ISO 23247. This study highlights the growing risks as digital twins become more autonomous and vulnerable to cyberattacks. The ISO 23247 standard addresses some basic security requirements such as confidentiality, authentication, and data integrity, but falls short in articulating more comprehensive functions—including availability, data traceability, auditing, non-repudiation, and governance frameworks--necessary to safeguard advanced digital twin-enabled AI simulation. For example, in our sample of primary studies, \textcite{liu2022digital} highlight the risk and challenge of data leakage and message tampering and proposed blockchain-based verification to ensure data integrity and traceability.

Thus, while the ISO 23247 reference architecture provides a good start for implementing AI-centered frameworks, such as DT4AI, it is somewhat limited in its current form in supporting AI simulation requirements. Key gaps include the lack of support for advanced data storage facilities, digital twin composition, ontological frameworks, credibility assessment (VVUQ), lifecycle management, and extra-functional properties, such as security.

\section{Conclusion}\label{sec:conclusion}

In this article, we analyzed the trends in digital twin-enabled AI simulation, and derived a conceptual reference framework to situate digital twins and AI with respect to each other. Our inquiry into the state of the art suggests that AI simulation by digital twins is a rapidly emerging field with demonstrated benefits in specific problem domains. At the same time, AI simulation by digital twins is still in its infancy, marked by limited usage of digital twin capabilities, simple lifecycle models, and lacking architectural guidelines---challenges that require active involvement from the digital twin engineering community. To foster involvement, we identify challenges and research opportunities for prospective researchers.

Our ongoing work is focusing on the architectural concerns of AI simulation. In the future, we plan to make strides towards standardization in collaboration with industry experts.

\newrefcontext[labelprefix=PS]
\printbibliography[keyword={primary},title={Primary studies},resetnumbers=true]

\newrefcontext
\printbibliography[notkeyword={primary},resetnumbers=true]

\end{document}